\pdfoutput=1

\documentclass[11pt]{article}

\usepackage[]{acl}

\usepackage{times}
\usepackage{latexsym}

\usepackage[T1]{fontenc}

\usepackage[utf8]{inputenc}

\usepackage{microtype}
\usepackage{adjustbox}

\usepackage{xspace}
\usepackage{times}
\usepackage{latexsym}
\usepackage{microtype}
\usepackage{natbib}
\usepackage{booktabs}
\usepackage{appendix}
\usepackage{tikz}
\usepackage{tikz-dependency}
\usepackage{amsmath}
\usepackage{amsfonts}
\usepackage{graphicx}
\usepackage{amsthm}
\usepackage{mathtools}
\usepackage{multirow}
\usepackage{url}
\usepackage{color}
\usepackage{subfig}
\usepackage{dsfont}
\usepackage{graphicx}
\usepackage{placeins}
\usepackage{float}
\usepackage{csquotes}
\usepackage{todonotes}

\usepackage[ruled,vlined,noend]{algorithm2e}

\usepackage[noabbrev,capitalize]{cleveref}
\crefformat{section}{#2\S#1#3}  

\newcommand{\control}{\textsc{Control Prefixes}\xspace}
\newcommand{\bartl}{BART$_{\text {LARGE }}$}
\newcommand{\bl}[1]{{\color{blue}  #1}}
\newcommand{\rd}[1]{{\color{red}  #1}}
\newcommand{\parhead}[1]{\medskip \noindent {\bfseries\boldmath\ignorespaces #1}\hskip 0.9em plus 0.3em minus 0.3em}

\title{\control for Parameter-Efficient Text Generation}

\author{Jordan Clive\\
  Imperial College London \\
   \And
  Kris Cao \\
  DeepMind, London, UK \\
  \And
  Marek Rei \\
  Imperial College London
 \AND \\
 \texttt{\{jordan.clive19,marek.rei\}@imperial.ac.uk} \\
 \texttt{kriscao@deepmind.com} 
 }

\begin{document}
\maketitle
\begin{abstract}

Prefix-tuning is a powerful lightweight technique for adapting a large pre-trained language model to a downstream application. However, it uses the same dataset-level tuned prompt for all examples in the dataset. We extend this idea and propose a dynamic method, \control, which allows for the inclusion of conditional input-dependent information, combining the benefits of prompt tuning and controlled generation. The method incorporates attribute-level learnable representations into different layers of a pre-trained transformer, allowing for the generated text to be guided in a particular direction. We provide a systematic evaluation of the technique and apply it to five datasets from the GEM benchmark for natural language generation (NLG). Although the aim is to develop a parameter-efficient model,  using only $0.1$–$3$\% trainable parameters, we show \control can even outperform full fine-tuning methods. We present state-of-the-art results on several data-to-text datasets, including WebNLG.

\end{abstract}

\section{Introduction}
Recently, approaches in text generation have been dominated by adapting one large-scale, pre-trained language model (PLM) to various downstream tasks. Such adaptation is often performed via fine-tuning, which necessitates updating and storing all of the parameters, resulting in multiple new language models (LMs), one for each task. This poses a considerable challenge to the deployment of NLP systems in practice, especially as the scale of PLMs continues to climb from millions to billions of parameters. Moreover, full fine-tuning has been shown to be unnecessarily profligate through overwriting natural language understanding (NLU) that could otherwise be shared among tasks \citep{poor_finetune}; it has also been shown that fine-tuned networks do not deviate substantially from the pre-trained one in parameter space \citep{aghajanyan20,dixit2020How}, implying the existence of parameter efficient alternatives.

Many researchers have sought to alleviate these issues by using \emph{fixed-LM} techniques, where the parameters of the base LM remain unchanged. An ever-growing subset of these methods can be considered prompt tuning, where language models are adapted to downstream tasks with the aid of a tuned prompt accompanying the input. A recent survey on prompt tuning \citep{prompt_survey}, however, notes the dearth of research exploring \emph{dynamic} prompts, which are input-dependent. This work fills this gap in the literature and considers such dynamic prompts. Existing controlled generation techniques either aim to generate text with specific target qualities, independent of overall task performance, or are methods that have the benefit of updating not only the attribute-level parameters but training all the parameters in the language model. 

We propose the \textit{dynamic} prompting method \control. The method extends prefix-tuning and integrates static task-specific prompts at every layer of a model, adding only $0.1$–$3$\% additional parameters to the base LM. With \control we aim to preserve the \emph{fixed-LM} property, while also allowing datapoint-specific attributes to act as guidance signals at the input-level. This is done by employing modular \emph{control prefixes}, which change alongside the input according to the guidance signal. Operating together with the static prompt parameters, these dynamic prompts can steer the frozen PLM to extend finer-grained control. The chosen attributes can provide additional information about the input, for example the domain of a data-to-text triple set, or it can specify some aspect of the desired output, such as the target length for text simplification. 

We evaluate our method on an array of text generation tasks, leveraging additional input-level information specific to each dataset. Our results show that our parameter efficient architecture outperforms previous approaches, many of them based on full fine-tuning, according to the WebNLG \citep{webnlg-2017}, DART \citep{dart} and E2E Clean \cite{E2Eclean} data-to-text datasets. In addition, our method attains higher human-assessed performance than existing systems for summarization on XSum \citep{xsum}. Although \control no longer operates in the standard setting for NLG tasks, by being not confined to just using the textual input, we focus on datasets where the attribute-level information is available as part of the task.

We also consider the common case where the attribute-level information is not available, and demonstrate that zero-shot learning with \control can be effective. We show similar control prefix representations are learned by the model for semantically similar attribute labels.

\begin{figure*}[ht]

    \centering
    \includegraphics[width=0.9\textwidth]{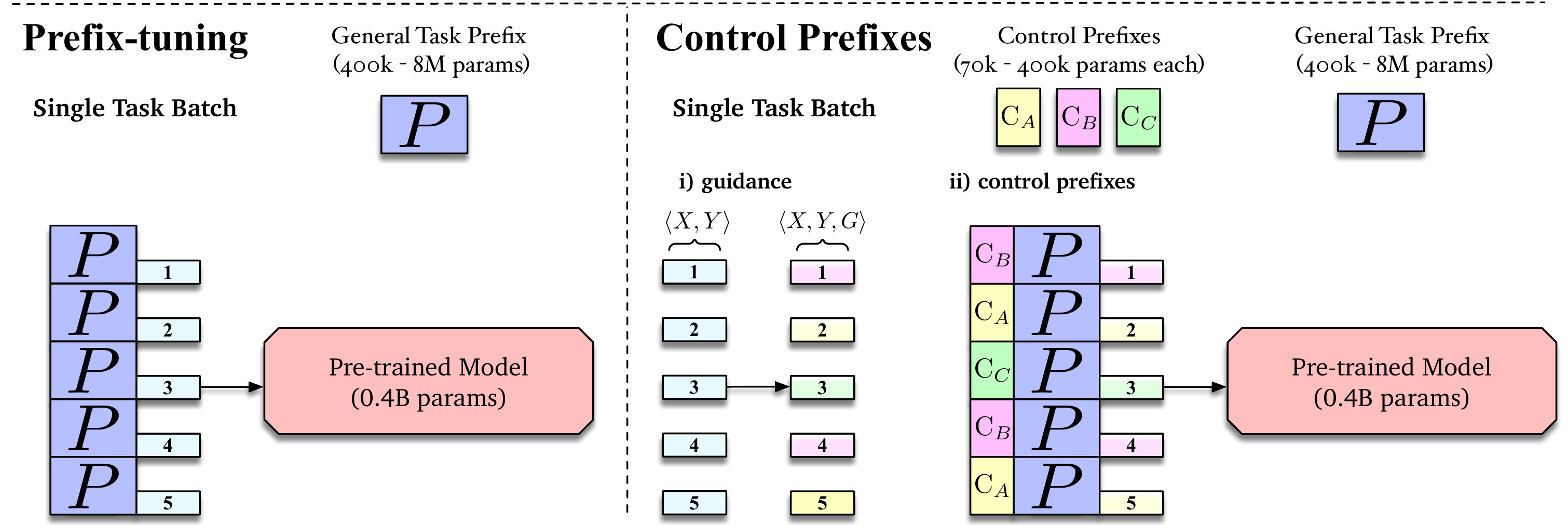}
    \caption{High-level diagram contrasting prefix-tuning and \control in the single-task setup for a PLM such as \bartl. The same single-task batch (examples 1,2,3,4 and 5) is considered for both setups. Left: Prefix-tuning has one general prefix $P$ for all examples. Right: \control utilizes additional attribute information at the input-level, $G$, in \textbf{i)}. This conditional information is used in \textbf{ii)} to dictate which control prefix ($C_A$, $C_B$, $C_C$) to use for a particular example in a batch. This takes advantage of prefix-tuning's capacity to include different prefixes in one forward pass.}
\end{figure*}

\section{Related Work}

\parhead{Prompt Tuning} Unlike the discrete text prompts used by GPT-3 \citep{brown2020gpt3}, in prompt tuning, soft prompts are learned through back-propagation to maximize the information from labelled data. This work focuses on tuning methods as zero-shot prompting performance lags far behind tuned models on supervised datasets \citep{lester}. Several successive works \citep{logs,ptune,lester} employ prompt-embedding tuning, which trains continuous embeddings prepended to the input embeddings. \citet{lisa} discovered that prefix-tuning was more effective than prompt-embedding tuning for text generation. In prefix-tuning, additional trainable key-value pairs, which are fixed across all examples, are used to augment the left context in every attention computation. Therefore, the prompt has constituents at every layer rather than being confined to steer the frozen LM only through the input as in embedding tuning.

\parhead{Controlled generation} A complementary field to prompt learning is controlled generation, which aims to incorporate various types of guidance (e.g. length specifications \citep{kikuchi} or highlighted phrases \citep{grangier-auli-2018-quickedit}) beyond the input text into the generation model. \citet{john_token} successfully trained a multilingual translation model with control tokens to encode each language. \citet{Keskar2019CTRL} pre-trained a 1.63B parameter model, also alongside conditional control tokens, and demonstrated these learnt to govern style, content, and task-specific behaviour. However, these models require the whole underlying LM to be fine-tuned alongside the control tokens for a particular task.

Alternatives exist, such as plug-and-play perturbations of the LM hidden states towards a target attribute \citep{pplm1,Dathathri2020Plug}. These methods use fixed LMs and are able to control target qualities such as sentiment and topic. However, they are slow at inference time due to requiring multiple passes for a single batch. The shift in conditional probability has also been shown to increase text degeneration \citep{zipf}.

\parhead{Dynamic prompts} There have been few works exploring \emph{dynamic} prompts \citep{prompt_survey,Tsimpoukelli}, which are input-dependent. Perhaps most similar to our work is work by \citet{att}, who use an attribute alignment function to form dynamic prompts. Unlike our work, the prompt does not have a static component and aims to generate text with specific target attributes, independent of task performance. With \control, the intention is to also maximize task-specific performance, which is why we maintain a large static prompt component to specify the task itself.

\section{\control}
\subsection{Background}
This work considers sequence-to-sequence tasks where the objective is to model the conditional probability $P(Y\mid X)$ with $X$ and $Y$ representing the tokenized input and output sequences respectively. For example, in summarization, $X$ could be an article and $Y$ would be a short target summary.

In this work we experiment with T5-large \citep{T5} and \bartl \citep{bart} as the underlying pre-trained LMs with parameters $\phi$; and as we consider fixed-LM methods, $\boldsymbol{\phi}$ \textbf{always remains frozen}. These models are Transformer encoder-decoder models where decoding proceeds auto-regressively. Let us denote $d$ to represent the hidden state dimension and $L$ the number of layers. We use $(E,Dc,Dm)$ to denote the three classes of attention present in each layer: self-attention in the encoder ($E$), decoder cross-attention ($Dc$) and decoder masked-attention ($Dm$). For an attention computation in the $l$-th layer, the query, key and value matrices are denoted $Q_{l} \in \mathbb{R}^{N \times d}$, and $K_{l}$, $V_{l} \in \mathbb{R}^{M \times d}$, where $N$ is the number of tokens in the series relating to queries, and $M$ is the number of tokens in the series relating to keys and values.

\subsection{Intuition}
Using a fixed PLM that captures broad natural language understanding provides the model with a parameter-efficient starting point which can be shared by many different tasks.
Combining this with a trainable task representation allows the model to learn information relevant to one particular task. Furthermore, introducing attribute-level parameters allows us to guide the generation into a required direction and provide the model with datapoint-level information. The general task-specific parameters can themselves adapt to the modular \emph{control prefixes}, which change according to the guidance signal for each input $X$. This demarcation of parameters enables fine-grained control to be extended to aid performance on downstream tasks. \control can therefore leverage input-level information while being a fixed-LM, parameter efficient method.\footnote{We use the term parameter efficient to denote methods adding $<$3\% additional parameters to a fixed LM's parameters.} For this work, we only consider discrete labels as attributes for the guidance signal.

\subsection{Description}
\label{sec:Description}

The model uses a general task prefix $P_\theta$ ("task-specific parameters") and also trains a set of control prefixes $C_\theta$ that change depending on the input  ("attribute-level parameters"). This requires attribute-level information or guidance $G$, to indicate which control prefixes to be used while processing a given input $X$.\footnote{We discuss cases where $G$ is not present in \cref{sec:zero}.} Let us consider the parallel corpus $\mathcal{Z}=\left\{\left\langle X^{j}, Y^{j},G^{j}\right\rangle\right\}_{j=1, . ., N}$, where $G^{j}$ indicates all the conditional attribute-level information for the sample $j$. The goal is to optimize through gradient descent the final inference parameters, $\theta$, whilst the underlying $\phi$ parameters of the pre-trained LM remain frozen:
\begin{equation}
   \theta^{*} = \arg \max _{\theta} \sum_{j=1}^{N} \log \; p \left(Y^{j} \mid X^{j}, G^{j} ; P_\theta,C_\theta,\phi\right).
\label{eq:control}
\end{equation}

\parhead{General Prefix} For each attention class $(E,Dc,Dm)$, a distinct prefix of key-value pairs is learnt, $P =  \{P_{1},\dots,P_{L}\}$, where $P_{l} \in \mathbb{R}^{\rho\times 2d} \; \forall l \in\{1, \ldots, L\}$. $P \in \mathbb{R}^{\rho\times 2dL}$ and $\rho$ is the prompt length, i.e. the number of additional key-value pairs in each attention computation.  In prefix-tuning\footnote{There has been confusion in recent work concerning different forms of prefix-tuning \citep{lisa}. For details and observations of the benefits (previously unremarked upon) conferred by key-value pair prefix-tuning, see Appendix \ref{app:prefix}.}, for an attention computation in the $l$-th layer, $K_{l}$ and $V_{l}$ are augmented to become
\begin{equation}
\begin{aligned}
K'_{l} = [P_{l,K};K_{l}]  \; , V'_{l} = [P_{l,V};V_{l}]
\label{eq:Pi}
\end{aligned}
\end{equation}
where $K'_{l},V'_{l} \in \mathbb{R}^{(\rho+M )\times d}$. The overall general prefix, parameterized by $\theta$, is $P_{\theta}=\left\{P^{E}, P^{D c}, P^{D m}\right\}$, where  $P_{\theta} \in \mathbb{R}^{\rho \times 6dL}$.

\parhead{Control Prefixes}
Let us consider one attribute with $R$ possible labels\footnote{The procedure can be generalized to multiple attributes; we use up to four attributes and varying control prompt lengths.}, such as the news domain of an article (e.g. sport, technology etc.), $C_{\theta}=\left\{C_{\theta, 1}, \ldots, C_{\theta, R}\right\}$, where  $C_{\theta, r} \in \mathbb{R}^{\rho_{c} \times 6 d L}$, $\forall r\in \{1 \ldots . R\}$. $C_{\theta, r}$ represents the control prefix learnt for the $r$-th attribute label and the parameter $\rho_{c}$ denotes the control prompt length for this \emph{particular} attribute. Let $\mathcal{A}$ be a function which returns the corresponding control prefix for the attribute label indicated by $G$. In \control the $K_{l}$ and $V_{l}$ are augmented to become
\begin{equation}
\begin{aligned}
  K_{l}^{\prime\prime}=\left[\mathcal{A}(G)_{l, K};P_{l, K} ; K_{l}\right], \\
  V_{l}^{\prime\prime}=\left[\mathcal{A}(G)_{l, V};P_{l, V} ; V_{l}\right]
\end{aligned}
\end{equation}
where $K_{l}^{\prime\prime}, V_{l}^{\prime\prime} \in \mathbb{R}^{(\rho_{c}+\rho+M) \times d}$.

\parhead{Shared Re-parameterization}
\label{sec:shared}
\citet{lisa} found that prefix optimization is stabilized by increasing the number of trainable parameters. This is achieved by introducing a feed-forward network to re-parameterize the prefix. Rather than one network, we use three distinct two-layered large feed-forward neural networks for each attention class, applied row-wise. For each attention class $(E,Dc,Dm)$, $P$ = $\text{MLP}(\tilde{P})$ where $\tilde{P} \in \mathbb{R}^{\rho \times d}$ is smaller than the matrix $P \in \mathbb{R}^{\rho\times 2dL}$, and each MLP has an intermediate dimension $k$ which we set to 800. The distinct MLPs and each $\tilde{P}$ are parameterized by training parameters $\tilde{\theta}$; thus, $\theta$ is a function of $\tilde{\theta}$ and $|\theta| < |\tilde{\theta}|$. Once training is complete, the final $\theta$ parameters can be saved for use at inference and the re-parameterization parameters dispensed with.

As described for the general prefix, $P_{\theta}$, each control prefix, $C_{\theta, r}$, comprises three constituents for each attention class: $C_{\theta, r}=\left\{C_{r}^{E},C_{r}^{Dc},C_{r}^{Dm}\right\}$. The re-parameterization of $C_{\theta, r}$ occurs in the same manner as $P_{\theta}$, sharing the same $\text{MLP}^{E}$, $\text{MLP}^{Dc}$ and $\text{MLP}^{Dm}$. When using a disjoint set of re-parameterizations for the control prefixes, learning becomes unstable and performance degrades.\footnote{This would also result in a significant increase in the number of training parameters $\tilde{\theta}$. In contrast, with the methodology outlined, each additional control prefix relates to only an additional $d\rho_{c}$ training parameters.}

Recent work by \citet{buhai2020empirical} show that over-parameterization can smooth
the optimization landscape. With this in mind, the three distinct re-parameterizations compel each prefix element to coordinate control for the particular attention class. For example, the rows of $P^{E}$ and $C_{r}^{E}$ lie in a vector space better coordinated for moderating the processing of the input sequence $X$ than $P^{Dm}$ and $C_{r}^{Dm}$. This is due to being formed from the shared mapping $\text{MLP}^{E}$.

\section{Experimental Setup}

\subsection{Datasets, Guidance and Metrics}

Examples of specific attribute labels for each task are found in the Appendix.\footnote{For data-to-text see Tables \ref{tab:app_web_qual}, \ref{tab:app_web20}; summarization see Table \ref{tab:app_qualitative_1} and simplification see Table \ref{tab:controlled_simp}.}

\parhead{Data-to-text}
The objective of data-to-text generation is to produce fluent text from structured input, such as a triple set (a set of subject-predicate-objects). Following \citet{lisa}, we evaluate on the data-to-text datasets DART \citep{dart} and WebNLG \citep{webnlg-2017}. 
However, we implement prefix-tuning for T5-large rather than GPT-2, as T5-large provides a stronger baseline and enables comparison with state-of-the-art (SOTA) systems.\footnote{\bartl exhibits inferior performance to T5-large on data-to-text; for example, 9.7 BLEU points lower on WebNLG Unseen \citep{ribeiro}.} We also report results on E2E Clean \citep{E2Eclean}, a dataset focused on the restaurant domain. 
We use the official evaluation scripts and report BLEU \citep{bleu_original}, METEOR \citep{lavie_meteor}, and TER \citep{ter} metrics.\footnote{Full results from the evaluation scripts, including machine-learned metrics can be found in Appendix \ref{app:addition}.}

WebNLG contains triple sets from DBPedia \citep{dbpedia}. The test set is divided into two partitions: ``Seen'', which contains 10 DBpedia categories present
in the training set, and ``Unseen'', which covers 5 categories never seen during training.\footnote{%
All the training category labels are visible in Appendix \ref{app:web2020}, where we visualize control prefixes corresponding to each training category.}
These categories, such as \emph{Airport} or \emph{Food} are used as a guidance signal in our experiments (indicated by $A_{1}$ in Table \ref{tab:datatotext1}); our approach for unseen categories is discussed in \cref{sec:zero}. 

Providing the category explicitly as guidance with \control may enable properties of triples belonging to a specific WebNLG category to be captured more effectively. This intuition is supported by studies showing a clear disparity in the performance of different model types between different categories  \citep{linearize1,web2020}. 
DART is an open-domain, multi-source corpus, with six sources: internal and external human annotation of both Wikipedia tables and WikiSQL, as well as the two existing datasets WebNLG and
E2E Clean. \citet{dart} showed fine-tuning T5-large on the WebNLG dataset with only the human annotated portion of DART achieves SOTA performance, whilst using the whole DART dataset is not as effective. Nevertheless, this inspired the idea of using the six DART sub-dataset sources as a controllable attribute, represented by $A_{2}$ in Table \ref{tab:datatotext1}. This strategy was inspired by previous work which incorporates auxiliary scaffold tasks \citep{Swayamdipta2018SyntacticSF,Dconan,tldr}.

\parhead{Simplification}
We use WikiLarge \citep{zhang2017sentence} as the training data  and evaluate on two simplification benchmarks: TurkCorpus \citep{turk} and ASSET \citep{asset}. Both benchmarks are composed of the same 2000 validation source and 359 test source sentences. However, the 10 ASSET references per source focus on a more diverse set of rewriting simplifications than the 8 TurkCorpus references per source.
\citet{MUSS} introduced \enquote*{\bartl with ACCESS}, which is a fine-tuned \bartl model trained alongside control tokens to condition on four simplification-specific attributes, such as the length compression ratio (the length of the target sequence relative to the source sequence).
We use the same controllable attributes in this work to directly compare with \citet{MUSS} (Table \ref{tab:simp_results}).
The control ratios are discretized into bins of fixed-width 0.05, capped to a maximum ratio of 2. At inference time, once the model has been trained with these oracle controls, the control ratios are set to desired values by tuning on the respective validation set. 

We report the non-learned metrics SARI \citep{turk} and FKGL \citep{fkgl}.\footnote{We use the FKGL and the latest version of SARI implemented in EASSE \citep{easse} which is used in \citet{MUSS}.} Unlike previous studies, we also use the machine-learned Q\&A metric QuestEval \citep{questeval} to assess our text simplification models.  

\parhead{Summarization}
As in \citet{lisa}, we report results on the XSum dataset \citep{xsum} using \bartl. XSum comprises
226,711 British Broadcasting Corporation (BBC) articles coupled with their single-sentence
summaries, where each sample corresponds to a unique URL. The URL contains information on whether the sub-directory is from the BBC Sport or BBC News page ($A_{1}$ in Table \ref{tab:main:experiments_gen}), and
further sub-directory information ($A_{2}$ in Table \ref{tab:main:experiments_gen}, where $A_2$ has 40 labels), for example (‘sport’, ‘formula1’) or (‘news’, ‘science’). The motivation for using this as guidance is that different sub-directories are likely to share
properties relating to how the information is presented; journalists are also usually confined to
one domain. We report on the customary ROUGE scores \citep{rouge_original}.

\subsection{Training Details}

For the data-to-text datasets, we follow \citet{ribeiro} and linearize the triples, prepending the special tokens <H>, <R>, and <T> before the subject, predicate, and object of an individual triple.\footnote{The embeddings relating to these special tokens are the only embeddings we
train, as our work is focused on fixed-LM methods.} We also prepend \enquote{translate Graph to English: } to every input \citep{T5}. Full training and hyperparameter details can be found in Appendix \ref{app:hyper}.

\section{Results}

\subsection{Data-to-Text}
Results in Table \ref{tab:datatotext1} show that for DART, both \control ($A_2$) and prefix-tuning attain higher
performance than the current SOTA, which is T5-large fined-tuned \citep{dart}, by 1.29 and 0.54 BLEU points respectively. This indicates \control can exert control over the frozen T5-large more effectively than prefix-tuning.

\begin{table*}[ht]
    \centering
    \resizebox{15cm}{!}{
    \begin{tabular}{l l r r r r r r r r r r r r}
    \toprule
         & & $\phi\%$ & \multicolumn{3}{c}{DART} &  $\phi\%$ & \multicolumn{3}{c}{WebNLG} &  $\phi\%$ & \multicolumn{2}{c}{E2E Clean} \\

         \cmidrule(lr){3-6} \cmidrule(lr){7-10} \cmidrule(lr){11-13}
               
         & & & BLEU & METEOR & TER $\downarrow$ & & \multicolumn{1}{c}{S} & \multicolumn{1}{c}{U} & \multicolumn{1}{c}{A} & & BLEU & METEOR \\
         \\
            & \quad T5-large fine-tuned & 100 & 50.66 & 40 & 43 & 100 & 64.89 & 54.01 & 59.95 & 100 & 41.83 & 38.1 \\
         & \quad SOTA & 100 & 50.66 & 40 & 43 & 100 & 65.82 & 56.01 &  61.44 & 100 & 43.6 & 39 \\
         \midrule
         & \quad Prefix-tuning & 1.0 & 51.20 & 40.62 & 43.13 & 1.0 & 66.95 & 55.39 & 61.73 & 1.0 & 43.66 & 39.0 \\
         & \quad \control ($A_1$) & - & - & - & - &  1.4 & \textbf{67.32} & 55.38 & 61.94 & - & - & -\\
         \midrule
         & \textbf{+Data: DART} \\
         & \quad Prefix-tuning  & 1.0 & 51.20 & 40.62 & 43.13 & 1.0 & 67.05 & 55.37 & 61.78 & 1.0 & 43.04 & 38.7 \\
     
            & \quad \control ($A_2$) & 1.1 & \textbf{51.95} & \textbf{41.07} & \textbf{42.75}  & 1.0 & 66.99 & 55.56 & 61.83 & 1.0 & \textbf{44.15} & \textbf{39.2} \\
         & \quad \control ($A_1$,$A_2$) & - & - & - & - & 1.4 & 67.15 & \textbf{56.41} & \textbf{62.27} & - & - & - \\
    \bottomrule
    \end{tabular} 
    }
    \caption{Data-to-text test set results reported on the respective official evaluation scripts. $\phi\%$ denotes the \% of additional parameters to the number of fixed-LM parameters required at inference time. T5-large fine-tuned results for WebNLG are from \citet{ribeiro} and for DART are from \citet{dart}. Note the results in the main body of the GEM paper \citep{gem} are reported on the validation set, rather than the test set as is done here. Several of the baseline results were only reported to the significant figures shown.  $A_{1}$ signifies models trained with control prefixes for the \emph{WebNLG category} attribute, and $A_{2}$ with control prefixes for the DART \emph{sub-dataset source} attribute. For WebNLG, S, U and A refer to BLEU scores for the \emph{Seen}, \emph{Unseen} and \emph{All} portions of the dataset. The DART results are reported on the official evaluation script for v1.1.1, the same version as the official leaderboard. A \control model attains state-of-the-art results for each dataset.}
    \label{tab:datatotext1}
\end{table*}

The SOTA for WebNLG is a T5-large model fine-tuned on WebNLG and the human annotated portion of DART \citep{dart}.\footnote{Additional training data is permitted by the organizers of the E2E Clean and WebNLG datasets.} Compared to this model, \control achieves a 0.83 higher BLEU overall, and 1.33 on the Seen categories. Notably, \control ($A_1$) outperforms
\control ($A_1$,$A_2$) on the Seen component of the dataset, but does not
generalize as well to the unseen categories. We argue that this illustrates the benefit of using both controllable attributes. The prefix-tuning model with additional DART data, like the SOTA, is trained on only the human annotated portion and yields a minor performance increase of 0.05 BLEU compared to prefix-tuning solely trained on WebNLG. We believe this indicates that for fine-tuning, training on a complementary type of additional data allows the PLM to maintain more NLU by not over-fitting a narrow distribution, leading to better LM generalization. In contrast, for prefix-tuning, much of this gain has already been realized by retaining the original frozen  parameters.

The SOTA \citep{dataturner} for E2E Clean consists of a fine-tuned GPT-2 with a semantic fidelity classifier trained on additional generated data. 
\control ($A_2$), which can leverage the heterogeneous DART datasets, outperforms this model in terms of the BLEU score.

\begin{table*}[t]
    \centering
    \resizebox{14cm}{!}{
    \begin{tabular}{l l r r r r r r r }
    \toprule
         & &  $\phi\%$ & \multicolumn{3}{c}{ASSET} & \multicolumn{3}{c}{TurkCorpus}  \\

         \cmidrule(lr){4-6} \cmidrule(lr){7-9} 
            
         & & & SARI & FKGL $\downarrow$ & QuestEval & SARI & FKGL $\downarrow$ & QuestEval\\
         \\
        
            & \quad Gold Reference &  - & 44.87 & 6.49  & 0.63$^{*}$  & 40.04  & 8.77 & 0.66$^{*}$ \\
   
         & \quad \bartl with ACCESS$^{\dagger}$ & 100 & 43.63 & 6.25 &  0.64$^{*}$ & 42.62 & 6.98 & 0.66$^{*}$ \\
        
        & \quad \bartl fine-tuned & 100 & 39.91$^{*}$  & 7.73$^{*}$ &  - & 39.55$^{*}$ & 7.73$^{*}$ & - \\
        \midrule
         & \quad Prefix-tuning & 1.8 & 40.12 & 7.28 & - & 39.06 & \textbf{7.28} & - \\ 
         & \quad \control & 1.8   & \textbf{43.58} & \textbf{5.97} & \textbf{0.64}  & \textbf{42.32} & 7.74 & \textbf{0.66} \\
    \bottomrule
    \end{tabular}
    }
    \caption{Simplification results on ASSET and TurkCorpus test sets. $^{\dagger}$This model is from \citet{MUSS}, where the authors fine-tuned \bartl model alongside control tokens for the four attributes. The \control model is trained with control prefixes for these same four attributes. Prefix-tuning and \control use \bartl as the fixed LM. The $^{*}$ denotes baseline results calculated in this study—the model outputs of \citet{MUSS} are publicly available. The \bartl with ACCESS and \control model are the average test set results over 5 random seeds. We bold the best results of parameter-efficient models in the results tables, while fully fine-tuned models and human performance are reported for reference.
}
    \label{tab:simp_results}
\end{table*}
\begin{table*}[ht]
    \centering
    \resizebox{\textwidth}{!}{
 \begin{tabular}{llccccccccc}
        \toprule
      
        & &  $\phi\%$ &
        \begin{tabular}{@{}c@{}}\textbf{Human} \\\textbf{overall}\end{tabular} & 
        \begin{tabular}{@{}c@{}}\textbf{Human} \\\textbf{conciseness}\end{tabular} &
        \begin{tabular}{@{}c@{}}\textbf{Human} \\\textbf{fluency}\end{tabular} & 
        \begin{tabular}{@{}c@{}}\textbf{Human} \\\textbf{no-hallucination}\end{tabular} & 
        \begin{tabular}{@{}c@{}}\textbf{Human} \\\textbf{informativeness}\end{tabular}
        & \multicolumn{3}{c}{\begin{tabular}{@{}c@{}}ROUGE \\R-1 \;\;\;\;\;\;  R-2 \;\;\;\;\;\; R-3\end{tabular}} \\
    \\

       & \bartl fine-tuned & $100$ &
$0.49_{-0.04}^{+0.03}$ & $0.50_{-0.03}^{+0.03}$ & $0.50_{-0.03}^{+0.03}$ & $0.52_{-0.03}^{+0.03}$ & $0.49_{-0.03}^{+0.03}$ & $45.14^{*}$ & $22.27^{*}$ & $37.25^{*}$ \vspace{2mm} \\ &
\vspace{2mm} 
        PEGASUS fine-tuned & 100 &
$0.49_{-0.03}^{+0.03}$ & $0.52_{-0.03}^{+0.02}$ & $0.49_{-0.02}^{+0.03}$ & $0.49_{-0.03}^{+0.03}$ & $0.49_{-0.03}^{+0.03}$
&$47.21^{*}$ & $24.56^{*}$ & $39.25^{*}$ \\
        &  T5 (11B) fine-tuned & 100 & $0.47_{-0.03}^{+0.03}$ & $0.49_{-0.02}^{+0.02}$ & $0.50_{-0.03}^{+0.03}$ & $0.49_{-0.03}^{+0.03}$ & $0.48_{-0.03}^{+0.03}$ & - & - & -	
 \\ 
         
\midrule
& {Prefix-tuning} & $3.0$ & - & - & - & - & - & $43.53$ & $20.66$ & $35.63$\\
& {\control ($A_{1},A_{2}$) } & $2.8$ & 
$\mathbf{0.51_{-0.03}^{+0.03}}$ & $\mathbf{0.53_{-0.02}^{+0.02}}$ & $\mathbf{0.51_{-0.03}^{+0.03}}$ & $\mathbf{0.53_{-0.03}^{+0.03}}$ & $\mathbf{0.49_{-0.03}^{+0.03}}$ & $\mathbf{43.81}$ & $\mathbf{20.84}$ & $\mathbf{35.81}$
\\

        \bottomrule
    \end{tabular}
    }
    \caption{Summarization results on XSum. The human-assessed results are from the GENIE benchmark, where the 95\% confidence intervals are computed with bootstrap re-sampling. Note the \bartl and PEGASUS fine-tuned results for the human-assessed dimensions are transcribed from \citet{GENIE}, whilst the automatic metric results, indicated by $^{*}$, are from \citet{bart} and \citet{peg}. Prefix-tuning and \control ($A_{1}$,$A_{2}$) use \bartl as the fixed LM. $A_{1}$ refers to the BBC news/sport page attribute and $A_{2}$ the further sub-directory attribute. We bold the best results of parameter-efficient models in the results tables for ROUGE, with fully fine-tuned models as reference. The public GENIE leaderboard is available at \url{https://leaderboard.allenai.org/genie-xsum/}.}
    \label{tab:main:experiments_gen}
    
\end{table*}

\subsection{Simplification}

Table \ref{tab:simp_results} reveals that prefix-tuning BART performs comparably to fine-tuning BART. When comparing our \control to fine-tuned ‘\bartl with ACCESS’ there is comparable performance in terms of SARI for ASSET, and better FKGL results on ASSET. For text simplification, \citet{MUSS} indicate the gains from using the controllable attributes, as assessed by SARI and FKGL, are mostly due to being able to calibrate the length ratio, with validation and test sets being drawn from the same distribution, as opposed to the WikiLarge training distribution. 
\control also achieves higher SARI and FKGL scores on TurkCorpus compared to the \emph{Gold Reference}, which evaluates against other human annotators.

\subsection{Summarization}

There is considerable inconsistency regarding author-conducted human evaluation for NLG \citep{vanderlee}. Therefore, we opted to submit our \control model outputs to an externally run evaluation framework, GENIE \citep{GENIE}, which provides an unbiased attestation of performance. Their sample size of 300 examples is larger than the 50 or 100 examples that have been previously used for XSum and is typical of human evaluation experiments \citep{xsum,dou_gsum_2021}.
Both human evaluation and automated ROUGE metrics can be seen in Table \ref{tab:main:experiments_gen}.
The confidence intervals indicate that this result is not necessarily definitive, but it also highlights that the quality of generations in this domain is not captured fully by ROUGE. For the datasets considered, the automatic metrics are the least reliable for XSum as it is the only dataset with a single gold reference. 

The results also show that \control performs better than prefix-tuning in terms of ROUGE. We are not able to report the same human-assessment results for prefix-tuning, as each participant of GENIE is limited to one submission and there is no existing result for prefix-tuning.

\section{Analysis}

\subsection{Visualizing Control Prefixes}
\label{sec:visual}
\begin{figure}[ht!]
    \centering
            
    \includegraphics[width=0.4\textwidth]{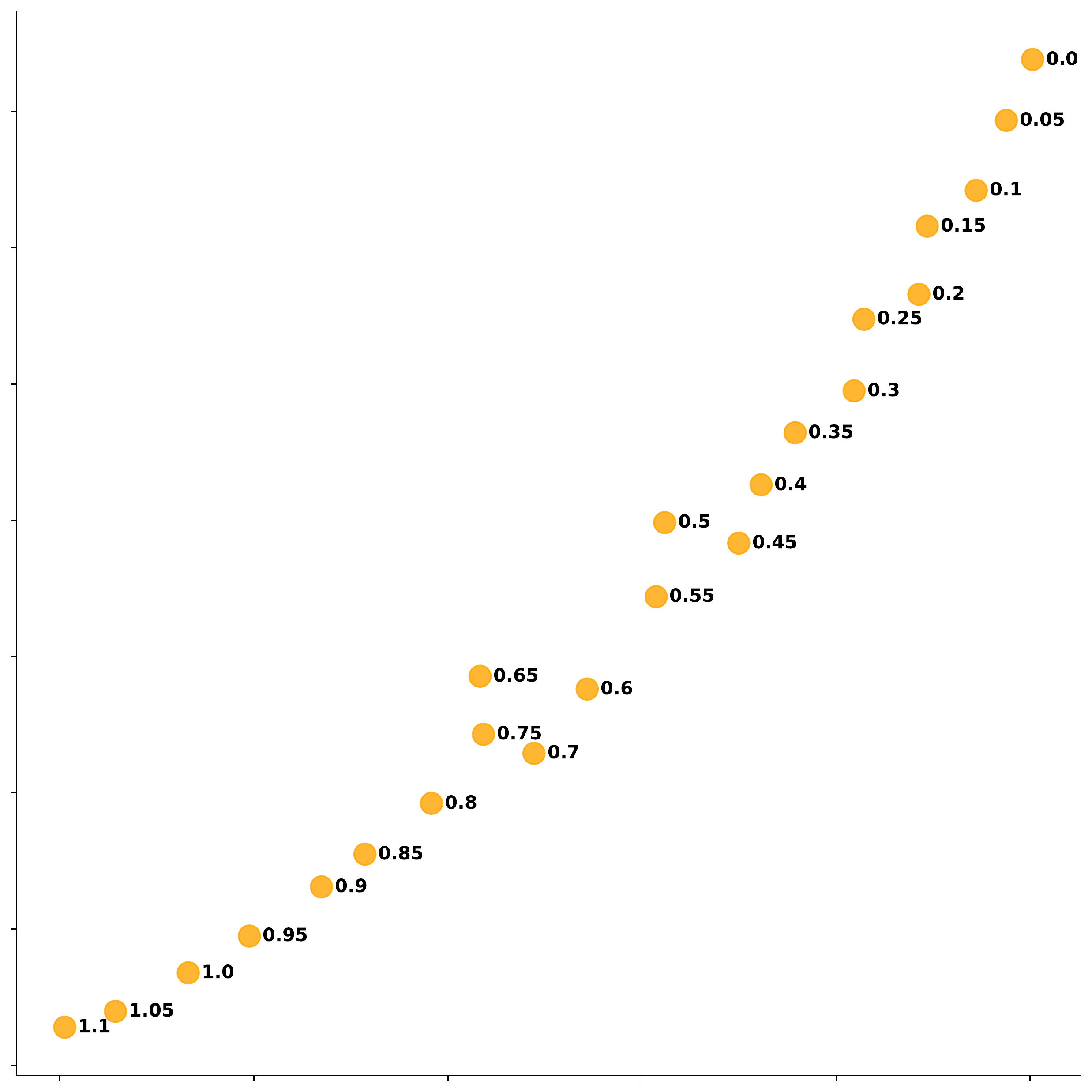}
    
    \caption{t-SNE visualizations for the decoder self-attention constituent of the simplification model's length compression control prefixes. Each circle represents a control prefix corresponding to each length ratio (bins of fixed width 0.05, from 0 to 1.1). 
    \label{fig:tsne}
    }
\end{figure}

Fig. \ref{fig:tsne} displays t-SNE \citep{tsne_orig} visualizations of the length compression control prefixes learnt as part of our simplification \control model.\footnote{A perplexity of 5 is used for all plots.} We plot only the decoder self-attention constituent of each control prefix (comprising multiple key-value pairs at each layer) as the length ratio directly concerns the target.\footnote{Plots for the encoder and decoder cross-attention constituents can be seen found in Appendix \ref{app:length_control}.} The relationship learnt by the control prefixes is very manifest, aided by the near uniform distribution of length ratios in the WikiLarge training dataset from 0 to 1.1. 

Fig. \ref{fig:tsne} establishes that for this simplistic attribute, different control prefixes corresponding to similar attribute labels (i.e. varying length ratios for the length attribute) share properties. Interestingly the decoder cross-attention of the control prefix is not as manifest. We believe this is due to \bartl being accustomed to the same cross-attention key-value pairs in each layer.

\subsection{Zero-shot Learning}
\label{sec:zero}
We argue that even for more complicated attributes, such as the WebNLG category attribute, if the attribute labels are semantically similar, the respective control prefixes will similarly assist the general task-specific prefix and the frozen LM during generation. 
Previous work has discussed the notion of task similarity \citep{achille} for prompt learning methods \citep{lester}; however, we argue prefixes concerning different labels of one attribute are more likely to overlap in terms of learnable properties than different tasks or whole datasets. 

In the case of WebNLG, where although no examples of the unseen category are present during training, a textual label for the category exists. These labels were available to all competition participants. This gives us some prior on the properties of the unseen
categories, which we show is enough to successfully zero-shot transfer with control prefixes. For each WebNLG model with the category attribute, we map each category's textual label, including for the unseen categories, to a Glove embedding\footnote{Glove Common Crawl (840B tokens, 2.2M vocab, cased, 300d vectors).} \citep{glove}. Then for each unseen category, we map to the seen category with the highest cosine similarity in embedding space, and use that control prefix at inference for the corresponding unseen sample. For example, the control prefix for the seen category \emph{SportsTeam} is used for examples relating to the unseen category \emph{Athlete}.\footnote{Appendix \ref{app:qual} displays model output for WebNLG along with the zero-shot procedure.}

Table \ref{tab:zero} shows a comparison of using an out-of-vocabulary (OOV) control prefix for each example with an unseen category, and the zero-shot transfer method for both WebNLG datasets\footnote{We also report results on WebNLG+ 2020 \citep{web2020}, the second official WebNLG competition, in Appendix \ref{app:web2020}.}. The OOV control prefix is trained on a random 2\% of the data for each accumulated batch. These results indicate that zero-shot transfer is more promising than a learned OOV representation. The result fundamentally depends on the WebNLG categories, and if similar textual labels pertain to similar triple sets that \control can utilize.

\subsection{Discussion}

We also investigated a simpler architecture \enquote*{prefix-tuning + control tokens} which informs the model of the identical guidance signal as in \control, but with trainable control tokens instead of control prefixes. Appendix \ref{app:control_tokens} reveals that \control consistently outperforms prefix-tuning + control tokens on the data-to-text and summarization datasets, while the results are both comparable to the \emph{Gold References} on simplification datasets. This indicates that \control is a superior parameter-efficient framework in leveraging additional information, whilst maintaining the \emph{fixed-LM} property.

The alternative method is less expressive than \control, by only exerting control through the embeddings rather than through each layer. \control fundamentally depends on the strength of the guidance signal and by adding the constraint of attribute information being available with the dataset the guidance signal is naturally weaker. However, we show that \control is a powerful general method which can utilize this signal to achieve a modest but consistent improvement across an array of tasks.

\begin{table}[t]
\centering
\resizebox{\columnwidth}{!}{%
\begin{tabular}{lccc}
\toprule
& \multicolumn{3}{c}{\textbf{Unseen Component}}
             \\ & \multicolumn{1}{c}{\# Examples}  & \multicolumn{1}{c}{\# Categories} & \multicolumn{1}{c}{BLEU}  \\ 

\emph{\textbf{WebNLG}}  & 891 & 5 &  \\ 

\quad OOV Representation &

  &  & 56.35 \\

\quad Zero-shot &
 &  & \textbf{56.41}  \\
\emph{\textbf{WebNLG+ 2020}} & 896 & 3 & \\

\quad OOV Representation &
  &  & 50.02 \\
\quad Zero-shot &
 &  & \textbf{50.39}  \\
\bottomrule
\end{tabular}%
}
\caption{A comparison of the performance on the \emph{Unseen} portions for WebNLG test sets, with i) a single OOV Control Prefix used for all samples from unseen categories, or ii) the zero-shot transfer approach outlined, utilizing the available textual labels.}
\label{tab:zero}
\end{table}

\section{Conclusion}

We introduce \control, a parameter-efficient controlled generation technique, which integrates a task-specific prompt alongside dynamic prompts to leverage additional input-level information. The method extends prefix-tuning, enabling the model to have finer-grained control over generated text, and assists in maximizing downstream task performance. 

We demonstrate that \control outperforms prefix-tuning and prefix-tuning with embedding level guidance, as well as existing approaches, on an
array of natural language generation tasks. Our method attains state-of-the-art results on several data-to-text datasets including WebNLG. This is despite learning <2\% additional parameters to the underlying LM parameters (which remain fixed).  
Additionally, our method holds the highest human evaluation ranking on the external platform GENIE for the summarization dataset XSum. 

\clearpage

\bibliographystyle{acl_natbib}
\bibliography{custom}

\clearpage

\appendix
\label{sec:appendix}

\section{Additional Results}

\label{app:addition}
Additional results using the official evaluation scripts for the data-to-text datasets are reported in Tables \ref{tab:dart_result_orig},\ref{tab:webnlg_17},\ref{tab:E2E} to supplement the results in Table \ref{tab:datatotext1}.

\section{WebNLG+ 2020 Results}
\label{app:web2020}

As NLG is notoriously challenging to evaluate, this work assesses model performance on five of the eleven datasets comprising GEM \citep{gem}, a benchmark that intends to provide robust datasets and 
reproducible standards across an array of NLG tasks. The GEM datasets used in this study are DART, E2E Clean, ASSET, TurkCorpus and WebNLG+ 2020. 

\begin{figure}[h]
    \centering
    \subfloat[\centering WebNLG]{\includegraphics[width=0.8\columnwidth]{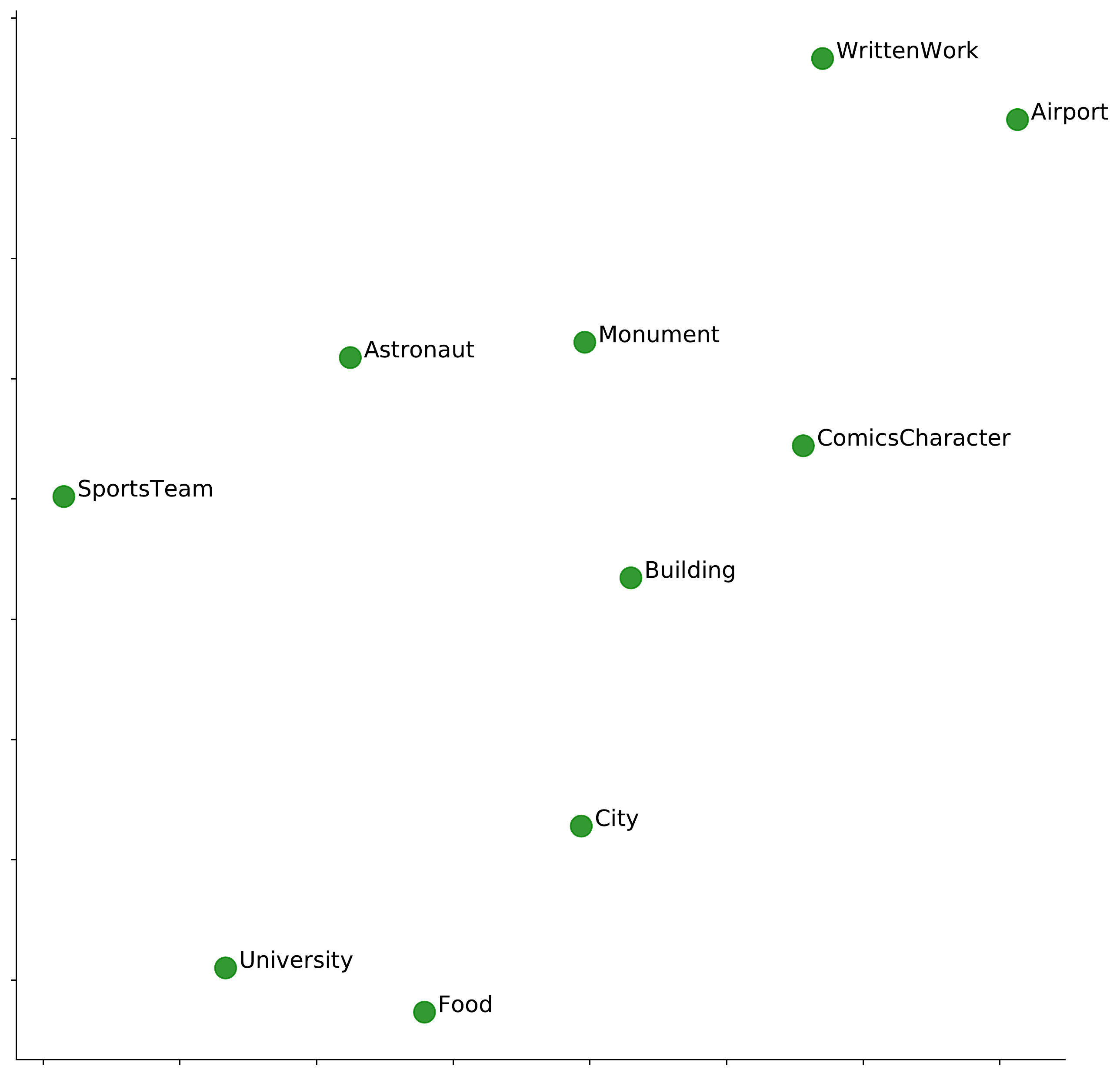} }%
    \qquad
    \subfloat[\centering WebNLG+ 2020 ]{\includegraphics[width=0.8\columnwidth]{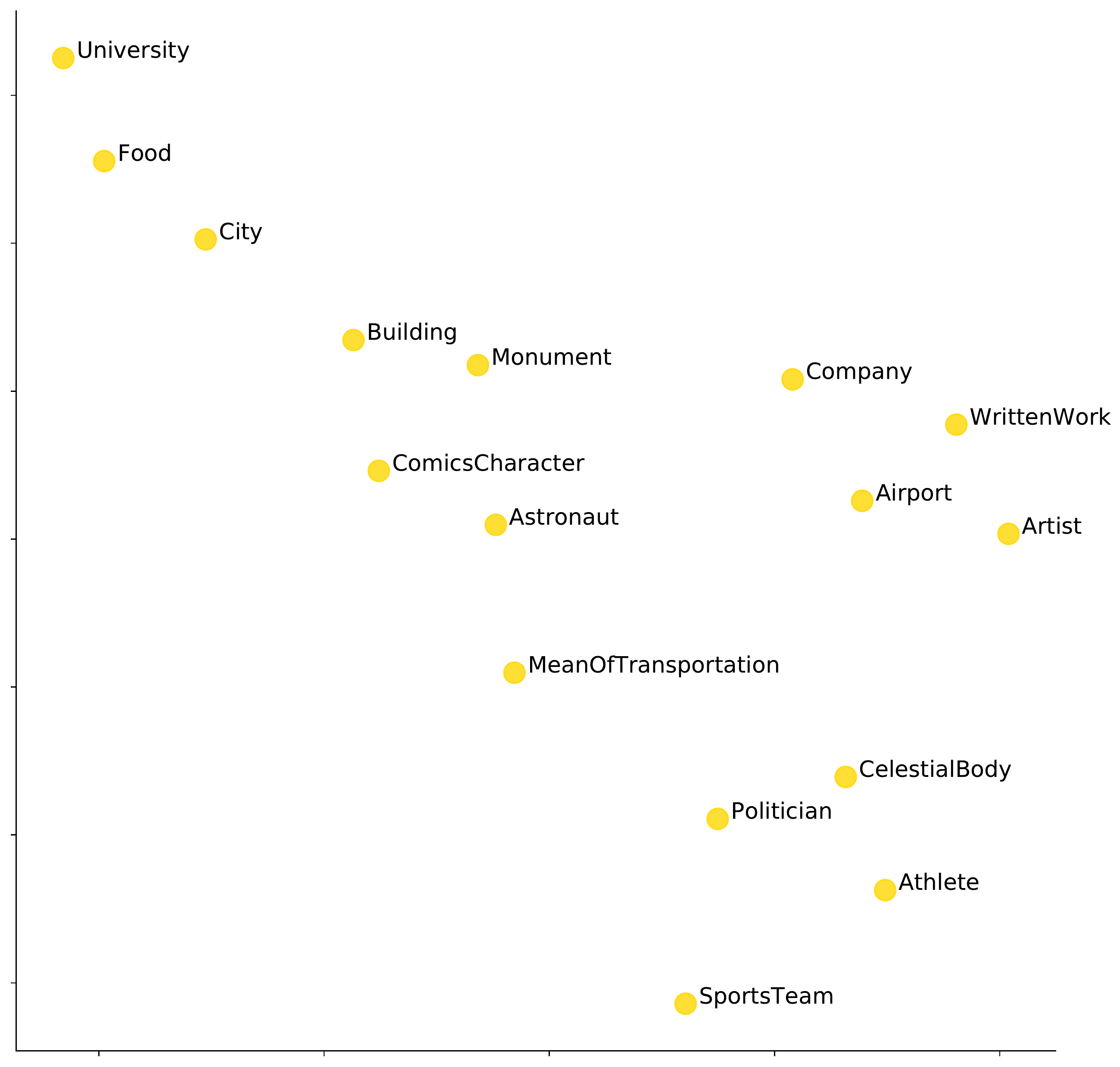} }%
    \caption{t-SNE visualizations for the encoder constituent of control prefixes representing WebNLG categories seen during training. Each circle represents a category seen during training for the \control ($A1$) model.  All 15 categories are seen categories in WebNLG+ 2020, along with the
category \emph{Company}. WebNLG+ 2020 has 3 additional unseen categories to those shown.}%
    \label{fig:web_cats_tsne}%
\end{figure}

WebNLG+ 2020  is not a component of DART—it was used for the second
official WebNLG competition \citep{web2020}. There are 16 training categories
(the 15 categories from WebNLG, but with new examples), alongside 3 unseen
categories. Table \ref{tab:webnlg2020} displays WebNLG+ 2020 results using the same model architectures as used for WebNLG. A similar pattern is revealed, in that \control outperforms prefix-tuning with \control ($A_1$,$A_2$) as the top-performing model. This illustrates again the benefit of using both controllable attributes.

In the WebNLG and WebNLG+ 2020 training sets, for the same tripleset, multiple distinct lexicalizations exist. In our experiments, the examples sharing identical tripleset inputs have the same triple order after linearization. This is to aid in comparison with current systems for WebNLG, DART and E2E Clean. Future work would have to assess if architecture-independent improvement in test-set performance can arise by random permutation of the order of triples for training set examples with identical tripleset inputs. The motivation being that this may improve the generalizability of the model, since the model would not learn the order of particular tripleset inputs.

\FloatBarrier

\section{Prefix-tuning}
\label{app:prefix}
We make two previously unremarked upon observations of the benefits conferred by using the key-value pair prefix-tuning described in \cref{sec:Description} compared to prefix-tuning involving augmenting the activations directly \citep{lora} or prompt-embedding tuning of prompt length $\rho$. i) The form discussed does not restrict the input length of the base LM. 
ii) The time complexity at inference time is reduced; for example, if we take a multi-head self-attention computation ($M=N$), the time complexity at inference time is $\mathcal{O}((N+\rho) N d + Nd^{2})$ rather than the greater $\mathcal{O}((N+\rho)^{2}d + (N+\rho)d^{2})$.

\section{Additional Training Details}
\label{app:hyper}

All implementations in this study are built on top of the Transformers library \citep{wolf}. As T5 has relative position biases, we set these in all layers pertaining to offsets where the key is part of a prefix to zero. 
For \bartl we adapt the original implementation \citep{lisa}.
Table \ref{tab:hyperparamters} displays the hyperparameters used when training the models reported in this paper.

The general prompt length and each control prompt length are
architecture-specific parameters that we choose based on performance on the validation
set.  We use gradient accumulation across batches to maintain an effective batch size above 64, a linear learning rate scheduler for all models and beam-search decoding. AdamW \citep{adamw}
and AdaFactor \citep{adafactor} were used for optimization. We chose the checkpoint with the highest validation score using BLEU for data-to-text, SARI for simplification and ROUGE-2 for summarization. For all tasks, we train our models on single Tesla V100-SXM2-16GB machines, with mixed precision for \bartl based models (fp16) and full precision for T5-large based models (fp32).

The \control models with the DART \emph{sub-dataset source} attribute ($A_{2}$) use DART as additional data and were trained in two stages: i) on DART, ii) solely on the downstream dataset. The WebNLG prefix-tuning model with DART data shown in Table \ref{tab:hyperparamters} uses only the human annotated portion of DART. The prefix-tuning models using all of the DART data for WebNLG and E2E Clean were similarly trained in two stages, with identical hyperparameters to \control models using $A_2$. Training prefix-tuning on all of DART for WebNLG yielded lower performance than with only the human-annotated DART portion as additional data, so was not reported in Table \ref{tab:datatotext1}.

Decoding specific parameters were not tuned—we instead mirrored what the top-performing fine-tuned based system used for the particular LM and dataset. For example, a beam width of 5 as in \citet{ribeiro} for T5-large on all data-to-text datasets. 

For XSum the source articles are truncated to 512 BPE tokens.


\section{Simplification Length Control}
\label{app:length_control}
\begin{figure}[h]
\centering
    \includegraphics[width=\columnwidth,scale=1]{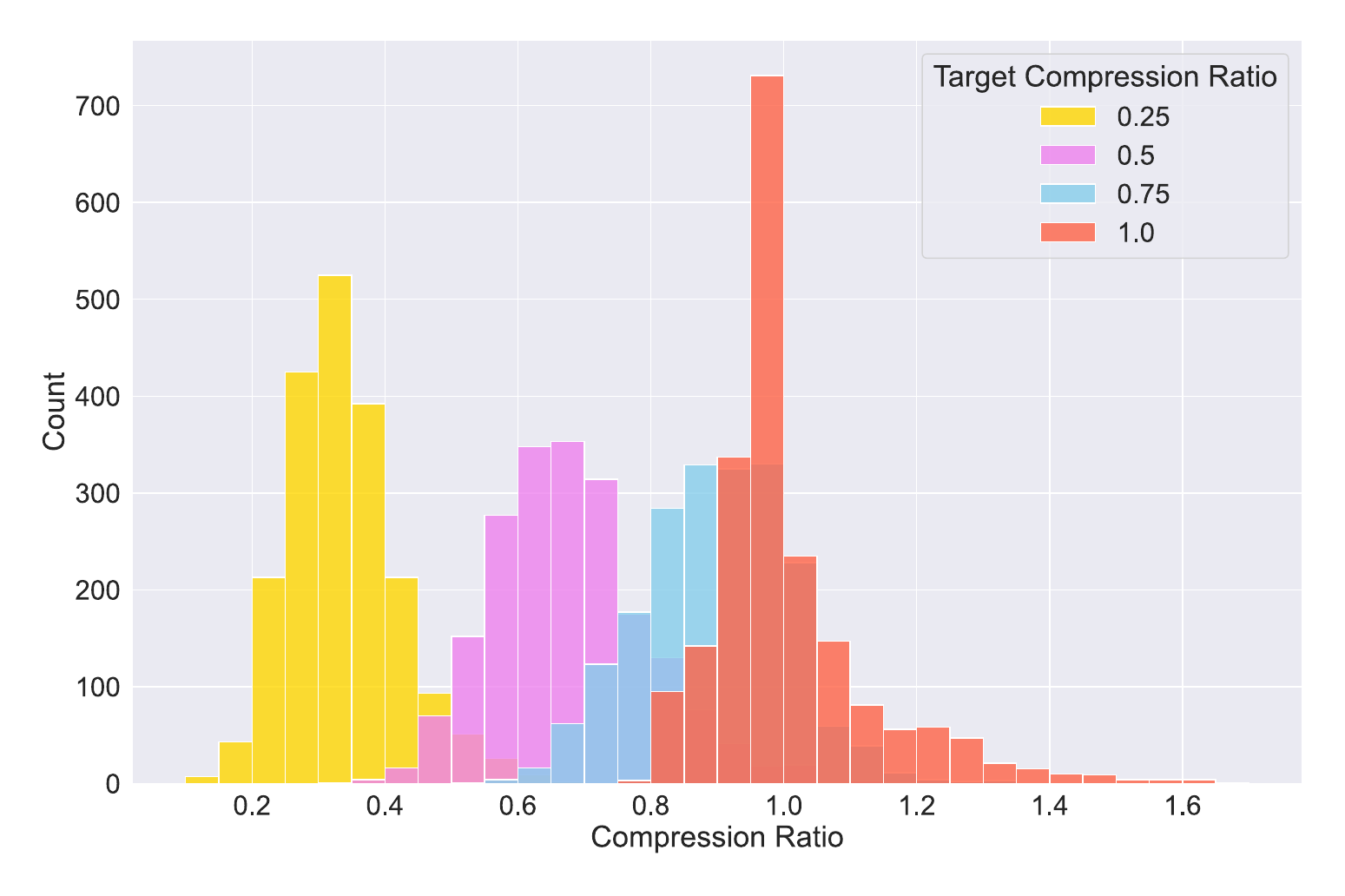}
        \caption{\small{Histogram illustrating the influence of different target length ratios on the actual length compression ratio output distribution for the simplification \control model on the TurkCorpus validation set.}}
\label{fig:app_hist}
\end{figure}

Fig. \ref{fig:app_hist} depicts the length compression ratio output distribution on the validation set for \control, where a length control prefix of a specific attribute value (0.25,0.5,0.75,1.0) is specified. This clearly demonstrates \control is capable of controlling the target length with respect to the input. Table \ref{tab:controlled_simp} displays example output generations with each of the 0.25,0.5,0.75,1.0 values specified.

\begin{figure}[h]
    \centering
    \subfloat[\centering Decoder Masked-attention ($Dm$)]{\includegraphics[width=0.98\columnwidth,scale=1]{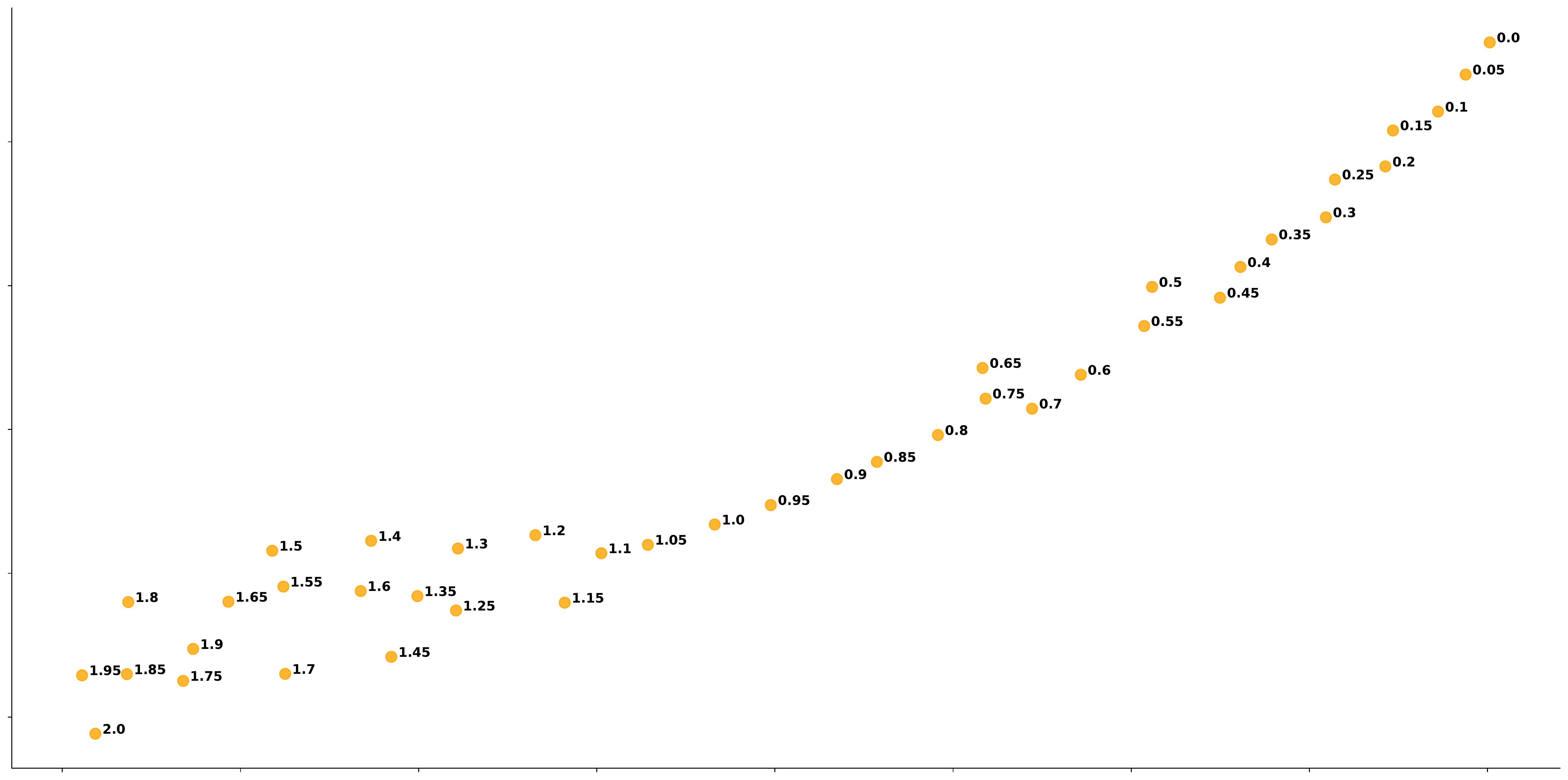} }%
    \qquad
    \subfloat[\centering Encoder ($E$)]{\includegraphics[width=0.98\columnwidth,scale=1]{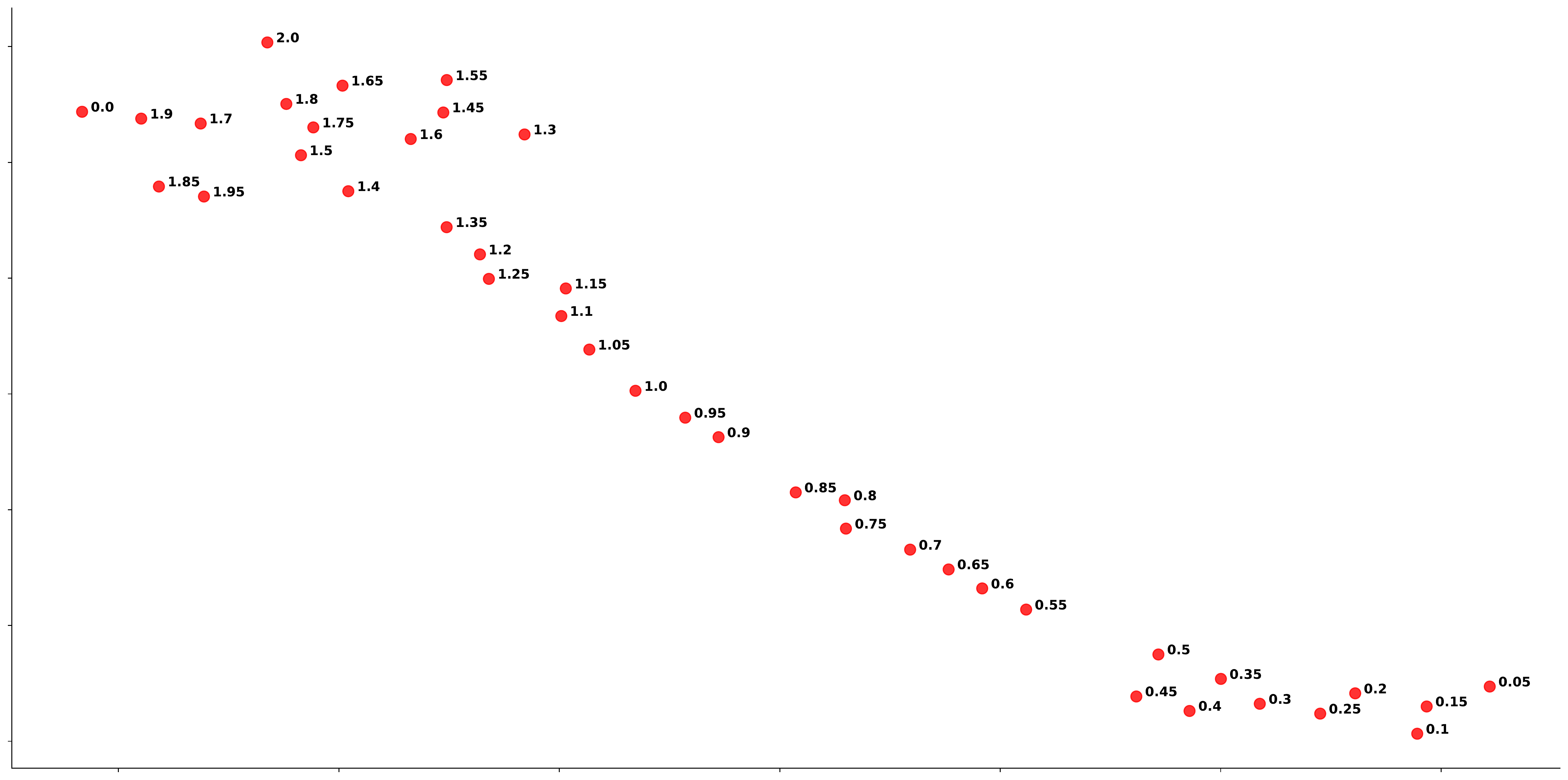} }%
    \qquad
    \subfloat[\centering Decoder Cross-attention ($Dc$)]{\includegraphics[width=0.98\columnwidth,scale=1]{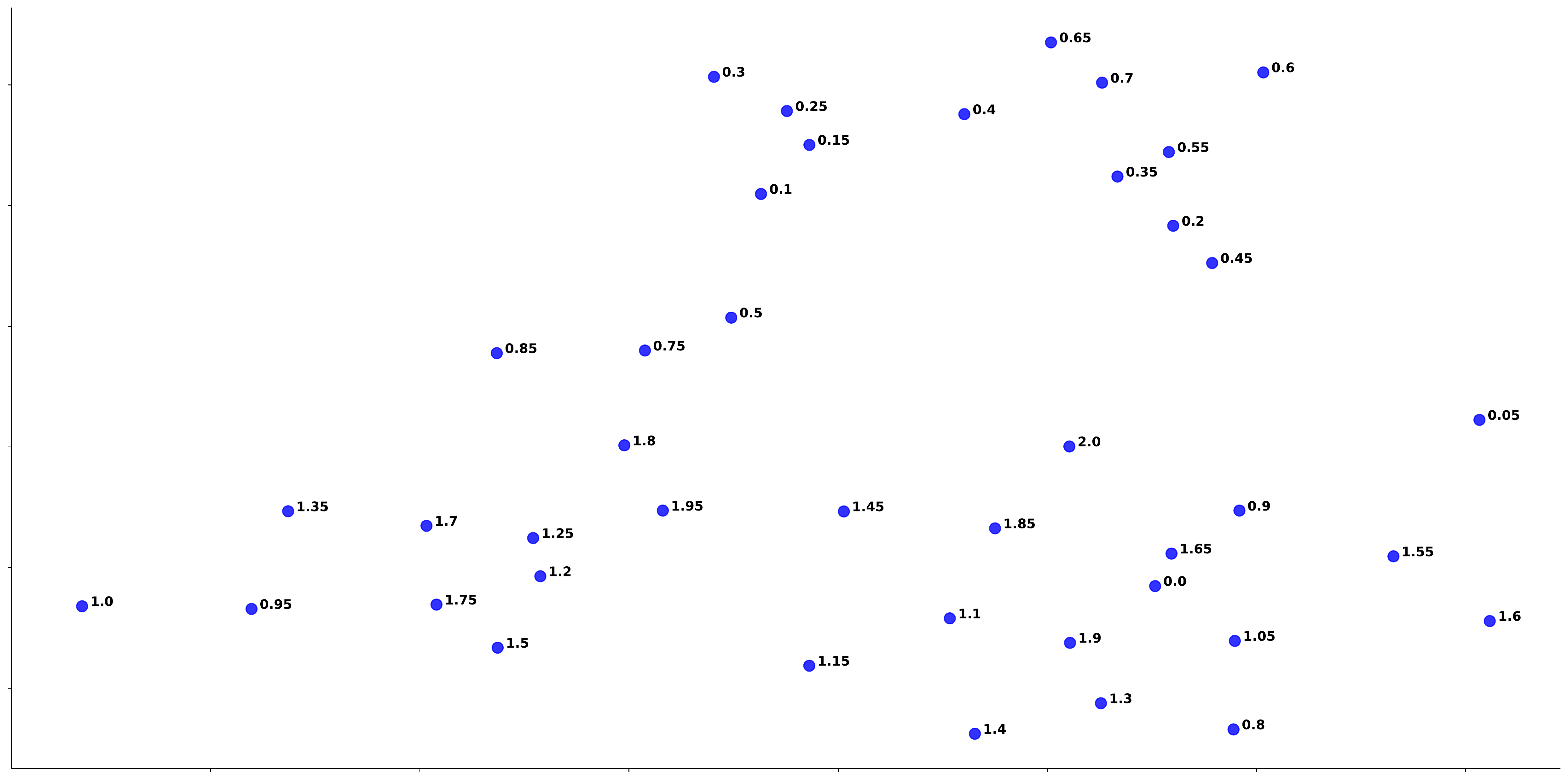} }%
 
    \caption{t-SNE visualizations for constituents of the length compression control prefixes learnt as part of the simplification \control model. Each diagram depicts representations of control prefixes corresponding to each length value (41 bins of fixed width 0.05, from 0 to 2) for a particular attention mechanism. The dimension represented on the x-axis is stretched from a 1:1 to 2:1 aspect ratio for labelling clarity.}%
    \label{fig:appen_tsne}%
\end{figure}

Fig. \ref{fig:appen_tsne} is supplementary to \cref{sec:visual}, showing all constituents of the length compression control prefixes for all attribute values. In the WikiLarge training data, there are far fewer training samples where the simplified output is much longer than the complex, original input in WikiLarge. This explains why the representations are not as interpretable for values greater than 1.2.

\subsection{QuestEval}
The \emph{Gold Reference} results for QuestEval\footnote{Although QuestEval can take references, the authors maintain that any improvement in correlation
with human performance is very minor.} are higher for TurkCorpus compared to ASSET in Table \ref{tab:simp_results}. We argue this is because the test set gold references are on average 114 characters for TurkCorpus, as opposed to 98 for ASSET. Therefore, the ASSET references contain less information to answer the generated queries during QuestEval evaluation; and thus, there is lower performance. We argue this shows a limitation with using QuestEval as a reference-less metric for simplification—by favouring longer generations.

\section{Prefix-tuning + Control Tokens}
\label{app:control_tokens}

We propose another architecture \enquote*{prefix-tuning + control tokens}, where all of the original LM parameters, $\phi$, still remain fixed, including the embedding matrix. Control has to be exerted through the few control embeddings and prefix-tuning's ability to steer the frozen $\phi$ parameters through $<2\%$ additional parameters.
We use this method to inform the model of the same discrete guidance information as in \control, but with control tokens instead of control prefixes.\footnote{Only the embeddings pertaining to the controllable attributes and the prefix are trained.} This alternative method is less expressive than \control, in much the same way as prefix-tuning is more expressive than prompt-embedding tuning. Prefix-tuning + control tokens also does not benefit from the shared re-parameterizations (\cref{sec:Description}) that we argue allow for more effective demarcation of control of the fixed LM in each attention class subspace. 

Table \ref{tab:control_tokens} reveals that \control outperforms prefix-tuning + control tokens on the data-to-text and summarization datasets, while the results are both comparable to the \emph{Gold References} on simplification datasets. This indicates that \control is better able to integrate and leverage guidance signal at the input-level, whilst maintaining the \emph{fixed-LM} property, than prefix-tuning + control tokens.

\section{Varying Prompt Length}
\label{app:prompt_length}

\begin{figure}[h]
    \centering
    \subfloat[\centering \bartl ]{\includegraphics[width=0.75\columnwidth]{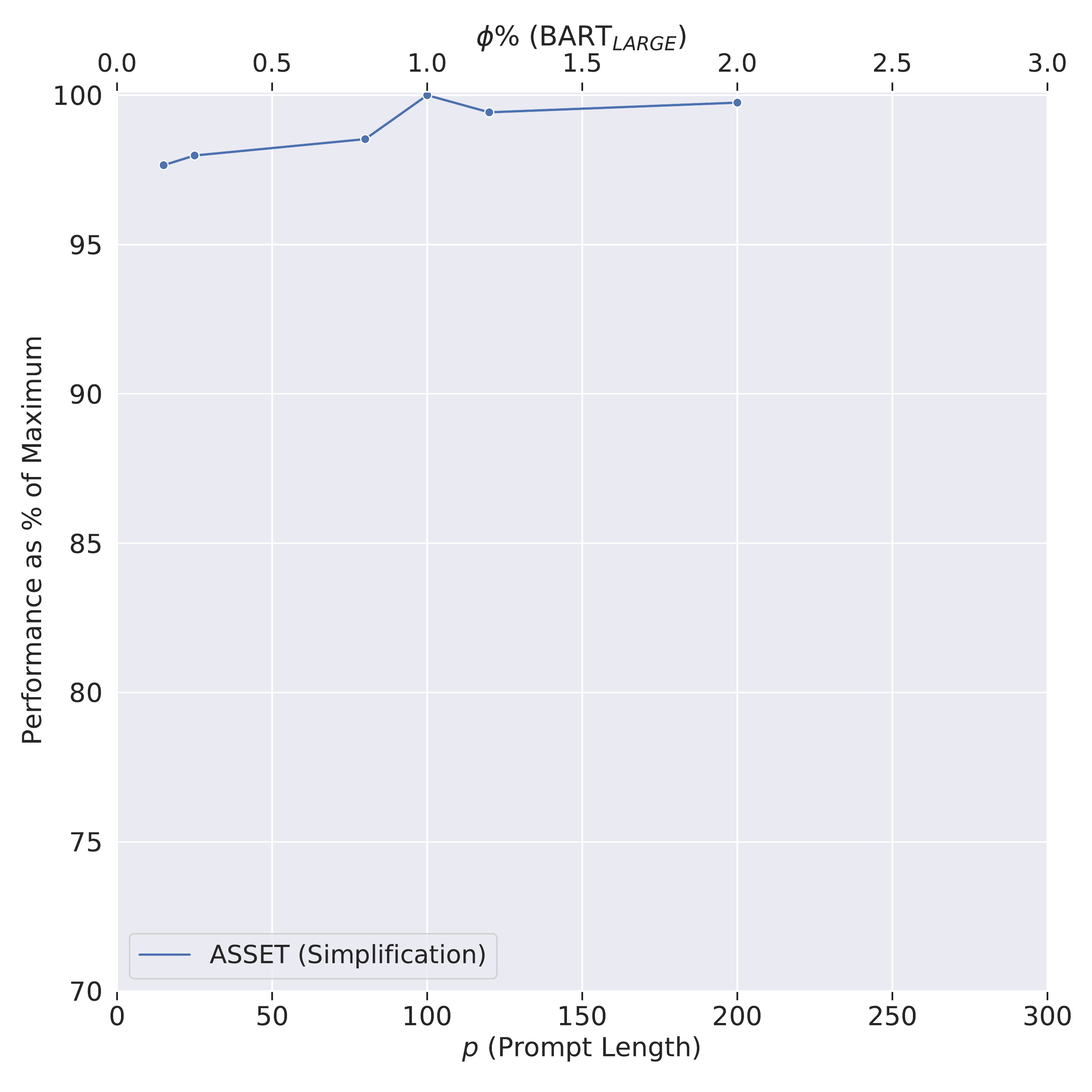} }%
    \qquad
    \subfloat[\centering T5-large ]{\includegraphics[width=0.75\columnwidth]{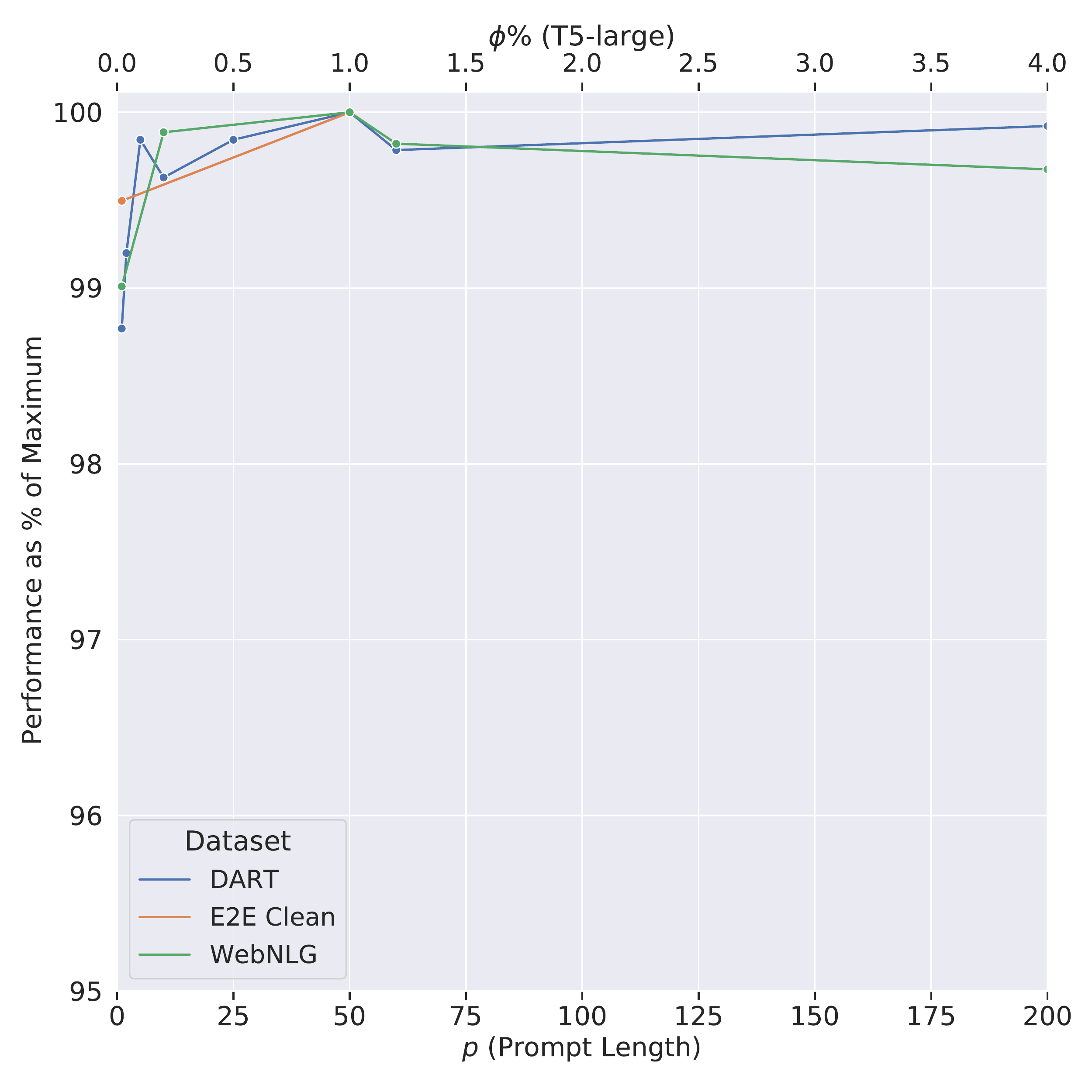} }%
    \qquad

    \caption{Prefix-tuning results of a model parameter search on several datasets for the optimal prompt length per dataset.  These results are for the metric monitored per task on the respective validation sets indicated in the legend. $\phi\%$ denotes the \% of additional parameters to the number of fixed-LM parameters required at inference time. The $y$-axis is a relative measure: the validation set performance as a \% of the maximum attained in the parameter search.} 
    \label{fig:prefix_len_proxy}%
\end{figure}

\FloatBarrier

Our research is not solely focused on parameter efficiency, but also on the effectiveness of adapting an already parameter efficient, fixed-LM method (adding <3\% additional parameters). The only way to add parameters with prefix-tuning is to increase the prompt length. XSum is the only dataset considered where performance does not plateau when increasing prompt length\footnote{We do not observe performance degradation, such as described by \citet{lora}, when utilizing different forms of prefix-tuning. This is shown in  \ref{app:prompt_length}.}, therefore we ensure \control does not have more parameters than prefix-tuning to ensure a fair comparison. 

The only way to add parameters with prefix-tuning is by increasing prompt length. Fig. \ref{fig:prefix_len_proxy} illustrates how performance saturation is observed—after a certain prompt length performance plateaus. Different datasets require varying prompt lengths to attain near maximum performance in a parameter search for prompt length. For the data-to-text datasets, near maximum performance (>99\% of the maximum validation score in the search) is reached with a prompt length of 1 or 2. 

\section{Qualitative Examples}
\label{app:qual}
For data-to-text, Table \ref{tab:app_web_qual} displays example \control output for WebNLG input belonging to unseen categories, along with the zero-shot procedure. Table \ref{tab:app_web_qual} depicts example \control ($A_{1}$,$A_{2}$) output alongside prefix-tuning model output for WebNLG+ 2020 input. For simplification, Table \ref{tab:app_simpcont} compares the fixed-LM guided generations of \control to the fine-tuned \bartl with ACCESS \citep{MUSS}. For summarization, Table \ref{tab:app_qualitative_1} depicts cherry-picked \control generated summaries for XSum input, alongside T5-large fine-tuned summaries that have higher ROUGE scores. This is to illustrate how \control can achieve higher human assessment through GENIE than top-performing fine-tuned models, whilst attaining lower automatic metric scores.


\FloatBarrier

\begin{table*}[htbp!]
\centering
\resizebox{0.7\textwidth}{!}{%
\begin{tabular}{lccccc}
\toprule
                             & $\phi\%$  & BLEU   & METEOR  & TER $\downarrow$   & BERTScore(F1)   \\ 
\\
\quad T5-large fine-tuned*               & 100 & 50.66 & 40 & 43 &  0.95 \\
\midrule 
\quad Prefix-tuning  & 1.0 & 51.20 & 40.62 & 43.13 & 0.95 \\
\quad \control ($A_{1}$) & 1.1 & \textbf{51.95} & \textbf{41.07} &  \textbf{42.75} & 0.95 \\
\bottomrule
\end{tabular}
}
\caption{Detailed results on the DART test set to complement Table \ref{tab:datatotext1}. T5-large fine-tuned is the current SOTA \citep{dart}. We report results on the official evaluation script for v1.1.1, the same version as the official leaderboard, available here: \url{https://github.com/Yale-LILY/dart}. *Results for this model were only reported to the significant figures shown. $\phi\%$ denotes the \% of additional parameters to the number of fixed-LM parameters required at inference time.
}
\label{tab:dart_result_orig}
\end{table*}

\begin{table*}[htbp!]
\centering
\resizebox{0.85\textwidth}{!}{%
\begin{tabular}{lcccccccccc}
\toprule
             & $\phi\%$ & \multicolumn{3}{c}{BLEU}  & \multicolumn{3}{c}{METEOR} & \multicolumn{3}{c}{TER $\downarrow$}  \\ 
             & &  S  & U & \textbf{A} & S  & U & \textbf{A} & S  & U & \textbf{A} \\
             
              \cmidrule(lr){3-5} \cmidrule(lr){6-8} \cmidrule(lr){9-11}\\
\quad T5-large  & 100  & 64.89 & 54.01 & 59.95 & 46 & 43 & 44 & 34 & 41 & 37 \\
\quad SOTA & 100 & 65.82 & 56.01 & 61.44 & 46 & 43 & 45 & 32 & 38 & 35 \\
\midrule 
\quad Prefix-tuning & 1.0  & 66.95 & 55.39 & 61.73 & 46.73 & 42.71 & 44.87 & 31.34 & 39.01 & 34.86 \\
\quad \control ($A_1$) & 1.4  & \textbf{67.32}  & 55.38		& 	61.94 & 46.78 & 42.77 & 44.92 & 30.96 & 39.01 & 34.65 \\
\midrule 
\small{\textbf{+Data: DART}} \\

\quad Prefix-tuning & 1.0 & 67.05 & 55.37 & 61.78 & 46.69 & 42.82 & 44.90 & 31.36 & 38.79 & 34.77 \\

\quad \control ($A_2$) & 1.0 & 66.99 & 55.56 & 61.83 & 46.67 & 42.87 & 44.91 & 31.37 & 38.53 & 34.65 \\
\quad \control ($A_1$,$A_2$)
 & 1.4  & 67.15 & \textbf{56.41} & \underline{\textbf{62.27}}  & 46.64 & 43.18 & 45.03 & 31.08 & 38.78 & 34.61 \\
\bottomrule
\end{tabular}%
}
\caption{Detailed results on the WebNLG test set to complement Table \ref{tab:datatotext1}. S, U and A refer to the \emph{Seen}, \emph{Unseen} and \emph{All} portions of the WebNLG dataset. Several of the baseline results were only reported to the significant figures shown.
}
\label{tab:webnlg_17}
\end{table*}

\begin{table*}[htbp!]
\centering
\resizebox{0.7\textwidth}{!}{%
\begin{tabular}{lcccccc}
\toprule
              & $\phi\%$  & BLEU  & NIST & METEOR & R-L & CIDEr      \\ 
\\
\quad T5-Large & 100 & 41.83
 & 6.41 & 0.381 & 56.0 & 1.97   \\
\quad SOTA & 100 & 43.6 & - & 0.39 & \textbf{57.5} & 2.0 \\
\midrule

\quad Prefix-tuning & 1.0 & 43.66 & 6.51 & 0.390 & 57.2 & 2.04  \\
\midrule
\small{\textbf{+Data: DART}} \\
\quad Prefix-tuning & 1.0 & 43.04 & 6.46 & 0.387 & 56.8 & 1.99  \\

 \quad \control ($A_2$) & 1.0 & \textbf{44.15} & \textbf{6.51} & \textbf{0.392} & 57.3 & \textbf{2.04} \\

\bottomrule
\end{tabular}%
}
\caption{Detailed results on the E2E Clean test set to complement Table \ref{tab:datatotext1}. The SOTA baseline result was only reported to the significant figures shown.
}
\label{tab:E2E}
\end{table*}

\begin{table*}[htbp!]
\centering
\resizebox{0.8\textwidth}{!}{
\begin{tabular}{lcccccc}
\toprule
               & $\phi\%$ &
                BLEU &	 METEOR  &	chrF++ &	TER $\downarrow$	 & BLEURT  \\ 
                \\

\quad T5-large*$^{\dagger}$ & 100  & 51.74  & 0.403 & 0.669 & 0.417  & 0.61 \\

\midrule 

 \quad Prefix-tuning  & 1.0 & 54.74 & 0.417 & 0.693 & 0.399 & 0.62 \\

\quad \control ($A_1$) & 1.6 &  54.97  & 0.417 & 0.693 & 0.398 & 0.62 	\\
\midrule
\small{\textbf{+Data: DART}} \\

\quad \control ($A_2$)   & 1.0 & 54.92  & 0.418 & 0.695 & 0.397 & 0.62 	\\

\quad \control ($A_1$,$A_2$)  & 1.6 & \textbf{55.41}  &	\textbf{0.419} &	\textbf{0.698} &	\textbf{0.392} &	\textbf{0.63} \\

\bottomrule
\end{tabular}%
}
\caption{\textbf{WebNLG+ 2020.} The overall WebNLG+ 2020 test set results using the official evaluation script. 
*As the model outputs are publicly available, we are able to run evaluation to achieve the same precision. $^{\dagger}$Results from \citet{nuig}, who before fine-tuning on the WebNLG+ data, further pre-train T5-large using a Mask Language Modelling objective (with 15\%
of the tokens masked) on the WebNLG corpus and a corpus of DBpedia. $A_{1}$ signifies models trained with control prefixes for the \emph{WebNLG category} attribute, and $A_{2}$ with control prefixes for the DART \emph{sub-dataset source} attribute. 
}
\label{tab:webnlg2020}
\end{table*}

\begin{table*}[t]
    \centering
    \resizebox{\textwidth}{!}{
    \begin{tabular}{l r r r r r r r r r}
    \toprule
         & & \multicolumn{1}{c}{DART} & \multicolumn{1}{c}{WebNLG} & \multicolumn{1}{c}{E2E Clean} & \multicolumn{2}{c}{ASSET} & \multicolumn{2}{c}{TurkCorpus} & \multicolumn{1}{c}{XSum} \\

         \cmidrule(lr){3-5} \cmidrule(lr){6-9} \cmidrule(lr){10-10}
               
         & & \multicolumn{3}{c}{BLEU} & \multicolumn{1}{c}{SARI} & \multicolumn{1}{c}{QuestEval} & \multicolumn{1}{c}{SARI} & \multicolumn{1}{c}{QuestEval} & R-2 \\
         \\
        
            & Prefix-tuning + Control Tokens & 51.72 & 61.89 & 43.57 & 43.64  & 0.63 & 42.36  & 0.66 & 20.70 \\
         & \control & 51.95 & 62.27 & 44.15 &  43.58 & 0.64  & 42.32 & 0.66 & 20.84 \\
   
    \bottomrule
    \end{tabular}
    }
    \caption{\textbf{Prefix-tuning + Control Tokens.} Comparison of our best \control model for each dataset with prefix-tuning + control tokens for the same attributes. The guided simplification models are the average test set results over 5 random seeds.}
    \label{tab:control_tokens}

\end{table*}

\begin{table*}[htb!]
\scriptsize
  \centering
  \begin{adjustbox}{max width=\textwidth}{
  \begin{tabular}{ccccccccccccc}

    Model &  Stage & L-rate & Opt & Warmup-steps & Epochs  & Batch Size & Effective Batch & Beam Width & LN-$\alpha$ & Min Target & Max Target & No Repeat Trigram\\\toprule
        \multicolumn{1}{c}{\it \textbf{DART (T5-large)}} \\
        \midrule
        \scriptsize{Prefix-tuning}   & - & 7e-5 & Ada & 2000  & 40 & 6 & 96 & 5 & 1 & 0 & 384 & No\\
    \scriptsize{\control ($A_{1}$)}   & - & 7e-5 & Ada & 2000  & 40 & 6 & 96 & 5 & 1 & 0 & 384 & No\\
    \midrule
    \multicolumn{1}{c}{\it \textbf{E2E Clean (T5-large)}}\\

\midrule
    \scriptsize{Prefix-tuning}   &  - & 8e-5 & Ada & 2000  & 50 & 6 & 96 & 5 & 1 & 0 & 384 & No\\
    \multirow{2}{*}{\scriptsize{\control ($A_{2}$)}} & 1 & 7e-5 & Ada & 2000  & 30 & 6 & 96 & 5 & 1 & 0 & 384 & No\\ & 2 & 5e-5 & Ada & 2000  & 50 & 6 & 96 & 5 & 1 & 0 & 384 & No\\
    
\midrule
    
    \multicolumn{1}{c}{\it \textbf{WebNLG (T5-large)}} \\
    \midrule

  \scriptsize{Prefix-tuning} & - & 7e-5 & Ada & 2000  & 30 & 6 & 96 & 5 & 1 & 0 & 384 & No\\
      \scriptsize{\control ($A_{1}$)}   & - & 7e-5 & Ada & 2000  & 40 & 6 & 96 & 5 & 1 & 0 & 384 & No\\
     \midrule
    \multicolumn{1}{l}{\scriptsize{\textbf{$\; \; \; \; \; \; \;$+Data: DART}}} \\

    \scriptsize{Prefix-tuning}   & - & 7e-5 & Ada & 2000  & 40 & 6 & 96 & 5 & 1 & 0 & 384 & No\\
    \multirow{2}{*}{\scriptsize{\control ($A_{2}$)}} & 1 & 7e-5 & Ada & 2000  & 30 & 6 & 96 & 5 & 1 & 0 & 384 & No\\ & 2 & 3e-5 & Ada & 2000  & 30 & 6 & 96 & 5 & 1 & 0 & 384 & No\\
      \multirow{2}{*}{\scriptsize{\control ($A_{1},A_{2}$)}} & 1 & 7e-5 & Ada & 2000  & 30 & 6 & 96 & 5 & 1 & 0 & 384 & No\\ & 2 & 3e-5 & Ada & 2000  & 30 & 6 & 96 & 5 & 1 & 0 & 384 & No\\

\midrule

\multicolumn{1}{c}{\it \textbf{XSum (\bartl)}}\\
\midrule
\scriptsize{Prefix-tuning} & - & 7e-5 & AdamW & 2000  & 40 & 8 & 128 & 6 & 1 & 10 & 60 & \checkmark \\
\scriptsize{\control ($A_{1},A_{2}$)}  & - & 7e-5 & AdamW & 2000  & 40 & 8 & 128 & 6 & 1 & 10 & 60 & \checkmark \\
\midrule
\multicolumn{1}{c}{\it \textbf{ASSET \& TurkCorpus (\bartl)}}\\
\midrule
\scriptsize{Prefix-tuning} & - & 5e-5 & AdamW & 2000  & 30 & 8 & 64 & 6 & 0.8 & 3 & 100 & \checkmark \\
\scriptsize{\control)} & - & 4e-5 & Ada & 5000  & 30 & 8 & 64 & 6 & 1 & 3 & 100 & \checkmark \\
\bottomrule
  \end{tabular}}
  \end{adjustbox}
\caption{\textbf{Hyperparameters.} Detailed hyperparameter reporting for the models in this work. If the training procedure is multi-stage, each stage is indicated. L-rate is the learning rate, all learning follows a linear learning rate scheduler; Opt refers to the optimizer, Ada (Adafactor) or AdamW; Effective Batch = Batch size x \# of gradient accumulation batches; LN-$\alpha$ refers to the $\alpha$ in length normalization during beam search.}
  \label{tab:hyperparamters}
\end{table*}

\clearpage

\begin{table*}[bth!]
\scriptsize
  \centering
  \renewcommand{\arraystretch}{1.3}
  \begin{tabular}{c|p{0.7\textwidth}}
  \hline
 
  \multicolumn{2}{c}{\textbf{ASSET Corpus}} \\ \hline
 & \textbf{Source:} The West Coast blues is a type of blues music characterized by jazz and jump blues influences, strong piano-dominated sounds and jazzy guitar solos, which originated from Texas blues players relocated to California in the 1940s. \\ \hline

Gold Reference$^\dagger$
& The West Coast blues has jazz and jump blues influences. It also has piano-dominated sounds and jazzy guitar solos, which originated from Texas blues players who moved to California in the 1940s. \\ \hline
\control  ($A_{1:4}$)\\Length Control: \rd{0.25}
& West Coast blues is a type of blues music from the United States.\\ \hline
\control  ($A_{1:4}$)\\Length Control: \rd{0.5}
& The West Coast blues is a type of blues music made from Texas blues players who moved to California in the 1940s. It has jazz and jump blues influences.\\ \hline
\control  ($A_{1:4}$)\\Length Control: \rd{0.75}
& The West Coast blues is a type of blues music that originated in Texas in the 1940s. It is characterized by jazz and jump blues influences, strong piano-dominated sounds and jazzy guitar solos.\\ \hline
\control  ($A_{1:4}$)\\Length Control: \rd{1.0}
& The West Coast blues is a type of blues music characterized by jazz and jump blues influences, strong piano-dominated sounds and jazzy guitar solos, which originated from Texas blues players who moved to California in the 1940s. 
\\ \midrule \hline

  \end{tabular}
  \caption{\textbf{Controlled text simplification}: a qualitative example from the ASSET Corpus validation set, with our corresponding \control model outputs. Only the control prefix for the length control ratio is varied, depicted in \rd{red}. $^\dagger$Note, this is one random gold reference out of 10.}
  \label{tab:controlled_simp}
\end{table*}

\begin{table*}[bth!]
\scriptsize
  \centering
  \renewcommand{\arraystretch}{1.3}
  \begin{tabular}{c|p{0.7\textwidth}}
  \hline
 
  \multicolumn{2}{c}{\textbf{ASSET Corpus}} \\ \hline
 & \textbf{Source:} The Great Dark Spot is thought to represent a hole in the methane cloud deck of Neptune. \\ \hline

Gold Reference$^\dagger$
& The Great Dark Spot represents a hole in the methane cloud of Neptune. \\ \hline
\control  \\
& It is thought that the Great Dark Spot is a hole in Neptune's methane cloud deck.\\ \hline
\bartl with ACCESS \\

& The Great Dark Spot looks like a hole in the methane cloud deck of Neptune.
\\ \midrule \hline

 & \textbf{Source:} Fives is a British sport believed to derive from the same origins as many racquet sports. \\ \hline

Gold Reference$^\dagger$
& Fives is a British sport developed from the same origins as many racquet sports. \\ \hline
\control \\

& Fives is a British sport. It is believed to have its origins in racquet sports.\\ \hline
\bartl with ACCESS \\

& Fives is a British sport. It is thought to come from the same as many racquet sports.
\\ \midrule \hline

 & \textbf{Source:} Nevertheless, Tagore emulated numerous styles, including craftwork from northern New Ireland, Haida carvings from the west coast of Canada (British Columbia), and woodcuts by Max Pechstein. \\ \hline

Gold Reference$^\dagger$
& Tagore copied many styles. These included craftwork from northern New Ireland, Haida carvings from western Canada and woodcuts by Max Pechstein. \\ \hline
\control\\

& Tagore emulated many different styles of art, including Haida carvings from the west coast of Canada (British Columbia), and woodcuts by Max Pechstein. \\ \hline
\bartl with ACCESS  \\

& Tagore copied many styles. He copied craftwork from northern New Ireland, Haida carvings from the west coast of Canada (British Columbia), and woodcuts by Max Pechstein.
\\ \midrule \hline

  \end{tabular}
  \caption{\textbf{\emph{Fixed-LM} vs fine-tuned controlled text simplification}. \control  and \bartl with ACCESS \citep{MUSS} generated simplifications chosen from the ASSET Corpus test set. $^\dagger$Note, this is one random gold reference out of 10 for each example. The examples shown for \control and \bartl with ACCESS are also randomly selected from one of the five model outputs.\label{tab:app_simpcont}}

\end{table*}

\begin{table*}[bth!]
\scriptsize
  \centering
  \renewcommand{\arraystretch}{1.1}
  \begin{tabular}{c|p{0.75\textwidth}}
  \hline
    \multicolumn{2}{c}{\textbf{WebNLG}} \\  \hline
    
 \\ Unseen Category: \bl{\textit{Athlete}}\\
  Zero-shot -> \rd{SportsTeam} & \textbf{Source:} <H> FC Torpedo Moscow <R> season <T> 2014-15 Russian Premier League <H> Aleksandr Chumakov <R> club <T> FC Torpedo Moscow <H> FC Torpedo Moscow <R> manager <T> Valery Petrakov <H> FC Torpedo Moscow <R> chairman <T> Aleksandr Tukmanov

  \\ \hline
Gold & Valery Petrakov is the manager of FC Torpedo Moscow and its chairman is Aleksandr Tukmanov. Aleksandr Chumakov plays for the club which spent the 2014-15 season in the Russian Premier League.\\ \hline
Prefix-tuning\\
  
& Aleksandr Tukmanov and Valery Petrakov are the managers of FC Torpedo Moscow. The club played in the Russian Premier League in 2014-15 and their chairman is Aleksandr Tukmanov. \\\hline
\control ($A_{1}$)\\
& Aleksandr Chumakov plays for FC Torpedo Moscow which is managed by Valery Petrakov. The club's chairman is Aleksandr Tukmanov and they played in the Russian Premier League in the 2014-15 season. \\ \midrule \hline

    \\ Unseen Category:\\ \bl{\textit{MeanOfTransportation}}\\
  Zero-shot -> \rd{Airport} & \textbf{Source:} <H> Costa Crociere <R> location <T> Genoa <H> Costa Crociere <R> parent Company <T> Carnival Corporation \& plc <H> AIDAstella <R> operator <T> AIDA Cruises <H> AIDAstella <R> builder <T> Meyer Werft <H> AIDAstella <R> owner <T> Costa Crociere

  \\ \hline
Gold & Carnival Corporation \& plc is the parent company of Costa Crociere in Genoa, who own the AIDAstella. AIDAstella was built by Meyer Werft and is operated by AIDA Cruises.\\ \hline
Prefix-tuning\\
  
& Costa Crociere is located in Genoa and is owned by Carnival Corporation \& plc. AIDAstella is operated by AIDA Cruises and was built by Meyer Werft. \\\hline
\control ($A_{1}$) \\
& Costa Crociere is located in Genoa and is owned by AIDA Cruises. AIDAstella was built by Meyer Werft and is operated by AIDA Cruises. The parent company of Costa Crociere is Carnival Corporation \& plc.
\\ \midrule 
\hline

  \end{tabular}
  \caption{\textbf{WebNLG example generations:} sources are shown in their linearized form, as fed to the T5-large based models, with prefix-tuning output and one of the gold references shown for comparison with \control output. Triplesets are from WebNLG unseen categories and the zero-shot procedure is depicted using the textual category labels. As an example, for the unseen category \emph{Athlete}, the closest Glove embedding belonging to a \emph{seen} category label in embedding space is \rd{SportsTeam}. Therefore the trained control prefix relating to \emph{SportsTeam} is used for this example at inference time.}
  \label{tab:app_web_qual}
\end{table*}

\begin{table*}[bth!]
\scriptsize
  \centering
  \renewcommand{\arraystretch}{1.1}
  \begin{tabular}{c|p{0.7\textwidth}}
  \hline
 
  \multicolumn{2}{c}{\textbf{WebNLG+ 2020}} \\ \hline
\bl{WebNLG} \rd{MeanOfTransportation}\\(Seen with Unseen Entities) & \textbf{Source:} <H> Pontiac Rageous <R> production Start Year <T> 1997 <H> Pontiac Rageous <R> assembly <T> Michigan <H> Pontiac Rageous <R> assembly <T> Detroit <H> Pontiac Rageous <R> production End Year <T> 1997 <H> Pontiac Rageous <R> body Style <T> Coupe <H> Pontiac Rageous <R> manufacturer <T> Pontiac \\ \hline

Gold
& The Pontiac Rageous was a car with a coupe body style manufactured by Pontiac. Assembled in both Michigan and Detroit, it went into production in 1997, ending in the same year.\\ \hline
Prefix-tuning

& The Pontiac Rageous is a coupe manufactured by Pontiac. It is assembled in Detroit, Michigan and began production in 1997.\\ \hline
\control ($A_{1}$,$A_{2}$)
& The Pontiac Rageous is manufactured by Pontiac in Detroit, Michigan. Its production began in 1997 and ended in 1997. The Pontiac Rageous has a coupe body style.

\\ \midrule \hline
  \bl{WebNLG} (Unseen) \\ Unseen Category: MusicalWork\\
  Zero-shot -> \rd{Artist} & \textbf{Source:} <H> Bootleg Series Volume 1: The Quine Tapes <R> genre <T> Rock music <H> Bootleg Series Volume 1: The Quine Tapes <R> preceded By <T> Squeeze The Velvet Underground album <H> Bootleg Series Volume 1: The Quine Tapes <R> record Label <T> Polydor Records <H> Bootleg Series Volume 1: The Quine Tapes <R> recorded In <T> San Francisco\\ \hline

Gold\\ & The Velvet Underground Squeeze album was succeeded by the rock album Bootleg Series Volume 1: The Quine Tapes, recorded under record label Polydor Records in San Francisco.\\ \hline
Prefix-tuning\\
  
& The record label of Bootleg Series Volume 1: The Quine Tapes is Polydor Records. It was recorded in San Francisco and was preceded by Squeeze The Velvet Underground. Its genre is rock music. \\\hline
\control ($A_{1}$,$A_{2}$)
& Squeeze The Velvet Underground was preceded by Bootleg Series Volume 1: The Quine Tapes, which was recorded in San Francisco and released by Polydor Records. The genre of the album is rock music. \\ \midrule \hline

  \end{tabular}
  \caption{\textbf{WebNLG+ 2020 generations:} sources are shown in their linearized form as fed to the T5-large based models. The DART sub-dataset \bl{Source} control prefix is highlighted, along with the final \rd{Category} control prefix. The zero-shot procedure is depicted for the Unseen Category \emph{MusicalWork}. The closest embedding belonging to a Seen category in embedding space is \rd{Artist}. }
  \label{tab:app_web20}
\end{table*}

\begin{table*}[bth!]
\scriptsize
  \centering
  \renewcommand{\arraystretch}{1}
  \begin{tabular}{c|p{0.75\textwidth}}
  \hline
 
  \multicolumn{2}{c}{\textbf{XSum}} \\ \hline
  \bl{news} \rd{world} & Kamal C Chavara was detained by the police in Kerala state on Sunday after the youth wing of the Hindu nationalist BJP lodged a complaint against him.
Last month, the Supreme Court ruled that the anthem must be played in every cinema before a film is screened.
Some 20 people have been held in Kerala and Tamil Nadu since then for remaining seated during the anthem.
Also, India's colonial-era sedition law has been often used against students, journalists, writers and social activists and those critical of the government.
Reports said that the BJP's youth wing lodged a complaint against a Facebook post by Mr Chavara which allegedly insulted the anthem. The post was apparently an excerpt from one of his books.
Senior police official Sateesh Bino told the NDTV news channel that the writer-activist "is being questioned for his controversial post on the national anthem on Facebook" and had been charged with sedition.
Earlier this month, 12 people were arrested at a cinema in Kerala, after they remained seated while the national anthem played.
The cinemagoers, who were attending an international film festival, were later freed but they face charges of "failure to obey an order issued by a public servant, thereby causing obstruction or annoyance to others".
And at a cinema in Chennai, eight people who did not stand for the anthem were assaulted and abused, police said. The eight were later charged with showing disrespect to the anthem. \\ \hline
Gold & A writer in India has been charged with sedition for allegedly showing disrespect to the national anthem. \\ \hline
T5-large fine-tuned\\(\textbf{70.97/48.28/70.97})
  
& A prominent Indian writer has been charged with sedition for defying the National Anthem.  \\ \hline
\control ($A_{1}$,$A_{2}$)\\(\textbf{59.46/34.29/54.05})
& An Indian writer-activist has been charged with sedition over a post on Facebook which allegedly insulted the national anthem. \\ \midrule \hline

\bl{sport} \rd{horse-racing} & The 33-1 shot, ridden by David Mullins and trained by Mouse Morris, triumphed at Aintree in April to become the first novice to win the race since 1958.
The nine-year-old, owned by the Gigginstown House Stud, has twice recovered from a cracked pelvis.
"We didn't want to send him back to Aintree with a big weight, that wouldn't be fair," said Gigginstown's racing manager Eddie O'Leary.
"He provided us with our first Grand National and we'll never forget him."
BBC horse racing correspondent Cornelius Lysaght:
"As the first Grand National winner for owner Michael O'Leary's burgeoning Gigginstown House Stud as well as the first novice chaser to win the race in nearly 60 years, Rule The World has his place in history.
"Though he ran highly respectably at Punchestown after Aintree, O'Leary had already hinted that, having defied serious injury to reach one of the great pinnacles, he had perhaps done his bit.
"What a season for Gigginstown, with success at Aintree, in the Irish National and Cheltenham Gold Cup, but at a price. Rule the World has been retired and there are doubts whether Gold Cup winner Don Cossack will race again."  \\ \hline
Gold & This year's Grand National winner Rule The World has been retired.\\ \hline
T5-large fine-tuned\\(\textbf{57.14/46.15/57.14})

& A Grand National-winning novice ridden by the brilliant rider Rule The World has been retired.  \\ \hline
\control ($A_{1}$,$A_{2}$) \\(\textbf{{55.17/44.44/55.17}})
&Winning Grand National hurdler Rule the World has been retired from racing at the age of nine. \\ \midrule \hline
  \end{tabular}
  \caption{\textbf{XSum generated summaries} for T5-large fine-tuned and \control based on \bartl. These are presented alongside the source document and the sole gold reference. Source documents are truncated to 300 words if necessary. \textbf{ROUGE-1/ROUGE-2/ROUGE-L} are reported in bold. The \bl{news/sport} control prefix and the related \rd{sub-directory} control prefix are highlighted.}
  \label{tab:app_qualitative_1}
\end{table*}

\end{document}